\documentclass[lettersize,journal]{IEEEtran}
\usepackage{amsmath,amsfonts}
\usepackage{algorithmic}
\usepackage{algorithm}
\usepackage{array}
\usepackage[caption=false,font=normalsize,labelfont=sf,textfont=sf]{subfig}
\usepackage{textcomp}
\usepackage{stfloats}
\usepackage{url}
\usepackage{verbatim}
\usepackage{graphicx}
\usepackage{cite}
\usepackage[colorlinks,
linkcolor=blue,
anchorcolor=blue,
citecolor=blue,backref]{hyperref}
\usepackage[all]{hypcap}
\usepackage{multirow}
\usepackage{ulem}
\usepackage{soul, color, xcolor}

\soulregister{\cite}7 
\soulregister{\citep}7 
\soulregister{\citet}7 
\soulregister{\ref}7 
\soulregister{\pageref}7 
\begin{document}
	
	\title{SeisFusion: Constrained Diffusion Model with Input Guidance for 3D Seismic Data Interpolation and Reconstruction}
	
	\author{Shuang Wang, Fei Deng, Peifan Jiang, Zishan Gong, Xiaolin Wei, and Yuqing Wang\vspace{-5mm}    
		\thanks{This work was supported by the National Natural Science Foundation of China under Grant 41930112. (Corresponding author: Fei Deng.)}
		\thanks{Shuang Wang, P. Jiang are with the Key Laboratory of Earth Exploration and Information Techniques of Education Ministry, College of Geophysics, Chengdu University of Technology, Chengdu 610059, China (e-mail: wangs@stu.cdut.edu.cn; jpeifan@qq.com).
			
			Fei Deng, Zishan Gong, Xiaolin Wei, and Yuqing Wang are with the College of Computer Science and Cyber Security, Chengdu University of Technology, Chengdu 610059, China (e-mail: dengfei@cdut.edu.cn; gzishan1119@qq.com; wxiaolin9@qq.com; wmmchang@qq.com).}}
	
	\markboth{Journal of \LaTeX\ Class Files,~Vol.~14, No.~8, August~2021}%
	{Shell \MakeLowercase{\textit{et al.}}: A Sample Article Using IEEEtran.cls for IEEE Journals}
	
	
	\maketitle
	
	\begin{abstract}
		Seismic data often suffer from missing traces, and traditional reconstruction methods are cumbersome in parameterization and struggle to handle large-scale missing data. While deep learning has shown powerful reconstruction capabilities, convolutional neural networks' point-to-point reconstruction may not fully cover the distribution of the entire dataset and may suffer performance degradation under complex missing patterns. In response to this challenge, we propose a novel diffusion model reconstruction framework tailored for 3D seismic data. To facilitate three-dimensional seismic data reconstruction using diffusion models, we introduce conditional constraints into the diffusion model, constraining the generated data of the diffusion model based on the input data to be reconstructed. We introduce a 3D neural network architecture into the diffusion model and refine the diffusion model's generation process by incorporating existing parts of the data into the generation process, resulting in reconstructions with higher consistency. Through ablation studies determining optimal parameter values, although the sampling time is longer, our method exhibits superior reconstruction accuracy when applied to both field datasets and synthetic datasets, effectively addressing a wide range of complex missing patterns. Our implementation is available at \hyperref[https://github.com/WAL-l/SeisFusion]{https://github.com/WAL-l/SeisFusion}.
	\end{abstract}
	
	
	\section{Introduction}
	\IEEEPARstart{s}{eismic} exploration extrapolates geological insights through the analysis of seismic data acquired from sensors\cite{dai2018seismic}. Nonetheless, the placement of sensors in specific locations is challenging due to geographical, physical, or economic constraints, leading to seismic data being gathered with varying degrees of missing information\cite{chen2019interpolation}, including consecutive and discrete missing traces. Consequently, the reconstruction of comprehensive seismic data emerges as a pivotal initial stage in the seismic data processing workflow.
	
	Presently, methods for reconstructing complete data can be broadly classified into two categories. The first category encompasses theory-driven traditional interpolation methods. For instance, prediction
	filter-based methods\cite{porsani1999seismic,gulunay2003seismic} utilize frequency or time domains for interpolation. Nevertheless, the selection of an interpolation window significantly impacts the results, posing considerable challenges in determining the optimal window\cite{wang2019deep}. Interpolation methods grounded in wave equations\cite{fomel2003seismic,ronen1987wave} leverage underground velocity as prior information to extrapolate and interpolate wave fields. However, they heavily depend on precise underground velocity models, which prove challenging in practical implementation. Sparse constraint methods\cite{latif2016efficient,wang2010seismic} transform seismic data via sparse transformations and employ sampling functions for interpolating missing data. Yet, this approach necessitates the selection of numerous empirical parameters, complicating the attainment of optimal outcomes\cite{niu2021seismic}. Additionally, there are low-order constraint methods based on compressive sensing\cite{zhang2019nonconvex,innocent2021robust}. however, these methods are unsuitable for handling continuous missing data. The second category comprises data-driven deep learning-based data reconstruction methods\cite{zhou2018seismic}. These methods capitalize on the robust feature extraction capabilities of convolutional neural networks and utilize upsampling techniques for reconstructing seismic data. Examples include Convolutional Autoencoder (CAE)\cite{wang2020seismic}, UNet\cite{park2019reconstruction}, and Generative Adversarial Network (GAN)\cite{dou2023mda} architectures.
	\begin{figure*}[!t]
		\centering
		\includegraphics[width=6in]{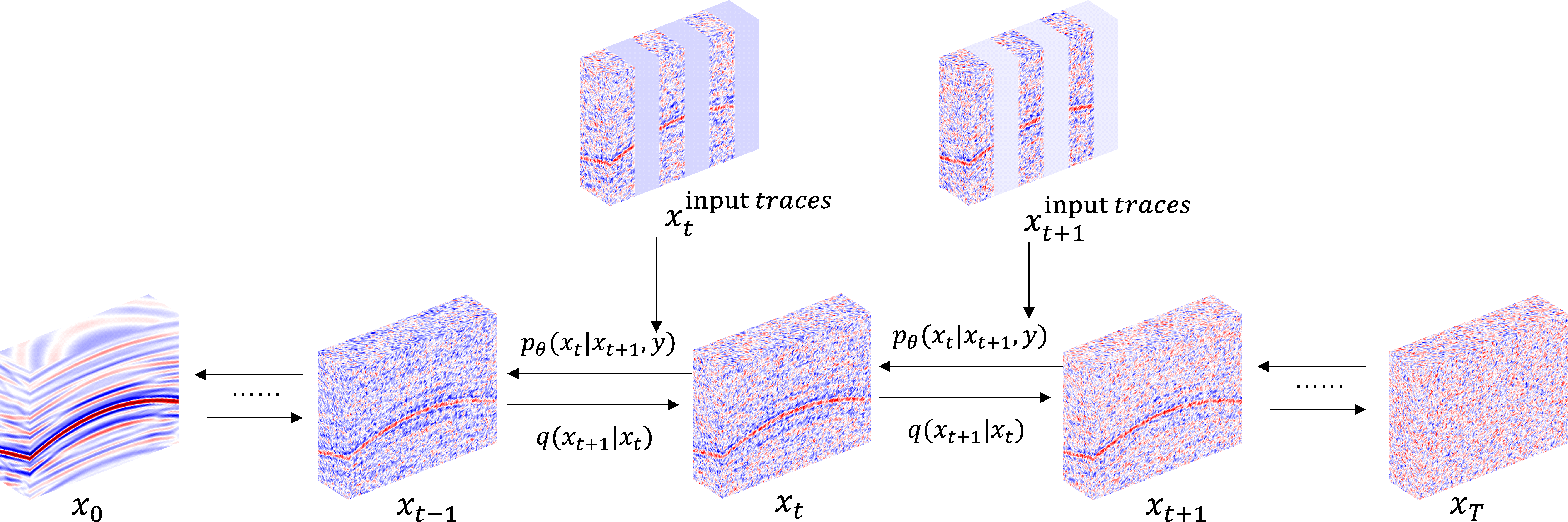}
		\caption{The diffusion model workflow involves $q$ for the forward encoding training process, where $x_0$ is gradually encoded into $x_T$. $p$ represents the reverse sampling generation process, where $x_0$ is gradually sampled from $x_T$, with $x_{t}^{input traces}$ serving as constraints added during the sampling process.}
		\label{condition}
	\end{figure*}
	
	Traditional methods, although theoretically capable of data interpolation and reconstruction\cite{abedi2022multidirectional}, heavily depend on manual parameter selection and struggle with interpolating large continuous missing data\cite{saad2023unsupervised}. In recent years, with the development of deep learning in recent years, the use of deep learning for seismic data processing has attracted a lot of attention~\cite{wang2022automatic}~\cite{jiang2023seismic}~\cite{mousavi2020earthquake}, and neural networks with different structures have been proposed for the reconstruction of seismic data\cite{pan2020partial,yu2021attention}. In the realm of 2D reconstruction, Wang \textit{et al.} \cite{wang2020seismic} advocated for using CAE to interpolate missing seismic data, while Chai \textit{et al.} \cite{chai2020deep} employed the UNet network for seismic data reconstruction. Abedi \textit{et al.}\cite{abedi2024ensemble} propose a new ensemble deep model and a customized self supervised training method for reconstructing seismic data with continuous missing traces. The proposed model consists of two U-net branches, each receiving training data from different data transformation modules. These transformation modules enhance underrepresented features and promote diversity among learners. Application to two benchmark synthetic datasets and two real datasets demonstrates that the accuracy of U-net is improved compared to traditional U-net methods. Inspired by the competitive results using the conditional diffusion probabilistic model(DPM) in the field of image superresolution, Liu and Ma \cite{liu2024generative} proposed a DPM-based seismic data interpolation method. The known parts of the data are used as constraints for the DPM to reconstruct the seismic data, demonstrating strong reconstruction performance. Deng \textit{et al.} \cite{deng2024seismic} introduced constraint modifications into the diffusion model and improved the sampling process by incorporating resampling techniques, achieving better results compared to classical convolution methods. Wang \textit{et al.} \cite{wang2024reconstructing} proposed a reconstruction method based on a classifier-guided conditional seismic denoising diffusion probabilistic model. This method employs classifier-guided techniques during diffusion model inference sampling, exhibiting very high reconstruction accuracy. However, since real seismic data is inherently 3D, focusing solely on inline slices for 2D reconstruction neglects crucial crossline information. One by one reconstruction of 2D slices can introduce discontinuities in stacked slices, particularly in datasets featuring large-scale or continuous missing data and other complex scenarios. Therefore, opting for 3D reconstruction is a more pragmatic approach. Chai \textit{et al.}\cite{chai2020deep3d} utilized an end-to-end 3D UNet for data reconstruction, observing superior performance compared to 2D methods. Qian \textit{et al.}\cite{qian2021dtae} proposed a deep tensor autoencoder, introducing tensor backpropagation, which exhibited commendable performance in reconstructing irregular data. Abedi and Pardo\cite{abedi2022multidirectional} proposed a multi-directional neural network that combines 2D and 3D learning capabilities. Initially, two networks are trained to perform simpler 2D reconstructions in the horizontal and vertical directions. Then, these networks are used as two parallel branches of a single network to perform the reconstruction of 3D data. Tests on synthetic data and field datasets show that this method achieves relatively accurate results and avoids the discontinuities associated with 2D U-Net. As GAN\cite{creswell2018generative} showcase potent generative capabilities, utilizing improved GAN networks for seismic data interpolation has gained traction. Dou \textit{et al.}\cite{dou2023mda} introduced a multidimensional adversarial generative adversarial network (MDA GAN), incorporating three discriminators and refining the loss function, resulting in robust reconstruction performance. Yu \textit{et al.}\cite{yu2023crossline} advocated for a CWGAN for data reconstruction, with results demonstrating its superiority over 3D UNet.
	
	Recent endeavors in 3D reconstruction predominantly rely on convolutional neural networks (CNNs)\cite{li2021survey}. However, CNNs encounter challenges in grasping the data distribution adequately, as they encode data into a feature space through convolution and then recovering the data from the feature space, resulting in a mere point-to-point mapping\cite{liu2024generative}. Regarding GAN networks, their training is difficult and unstable\cite{dhariwal2021diffusion}. To train a good GAN network model, proper initialization and setting of empirical parameters are required. Additionally, GAN networks still fall within the realm of convolutional neural networks and represent a point-to-point reconstruction approach\cite{wang2017generative}. Convolutional neural networks cannot cover the entire distribution of the dataset \cite{dhariwal2021diffusion}. When the data to be reconstructed exhibits other complex missing patterns, these convolutional neural networks may experience varying degrees of performance degradation.
	
	Diffusion models\cite{croitoru2023diffusion}, a type of generative model, have recently garnered significant attention for their remarkable success\cite{cao2024survey}. Unlike convolutional neural networks, diffusion models learn the probability distribution of target data\cite{yang2023diffusion}, offering desirable attributes such as comprehensive distribution coverage, fixed training objectives, and scalability \cite{nichol2021improved}. Consequently, diffusion models exhibit superior generative performance\cite{luo2022understanding}. As seismic data reconstruction transitions from 2D to 3D, the data's complexity increases exponentially. The distribution coverage feature of diffusion models enables them to comprehensively capture the distribution of 3D seismic data, resulting in enhanced performance. Therefore, diffusion models prove more adept for 3D seismic data reconstruction tasks. Nonetheless, direct utilization of diffusion models for 3D reconstruction poses challenges. Primarily, diffusion models are unconditional generative models\cite{song2020denoising}, capable of sampling and generating data resembling the probability distribution of the training set without constraints on the generated outcomes. Consequently, utilizing diffusion models directly for reconstruction may yield reconstructed results conflicting with the input data, which is undesirable for reconstruction tasks. Moreover, existing diffusion models designed for image generation are 2D generative models\cite{zhang2023text} lacking 3D generative capabilities. To address these issues and adapt diffusion models for 3D seismic data reconstruction, we introduce a 3D constrained diffusion model with input guidance (SeisFusion). By learning the probability distribution of seismic data and utilizing input data to guide and constrain the sampling process, SeisFusion effectively accomplishes reconstruction tasks, demonstrating superior reconstruction accuracy in model experiments compared to existing methods. The contributions of this paper are outlined below:
	
	1. Introduced a conditional supervision constraint into the diffusion model, using the known parts of the data as constraints to guide the generated data of the diffusion model, thereby ensuring higher consistency with the known data.
	
	2. Introduced a 3D neural network architecture into the diffusion model, successfully extending the 2D diffusion model to 3D space, and introduced a diffusion model-based reconstruction model (SeisFusion), into 3D seismic data reconstruction work.
	
	3. Improved the generation and reconstruction process of the model by guiding the diffusion model’s sampling process with the known parts of the data. This approach ensures that the sampling results include prior information from the known data, enabling the generation of reconstruction data with higher consistency.

	\section{Diffusion model}\label{ddpm}
	Diffusion models, currently the most advanced generative models\cite{rombach2022high}, acquire the probability distribution of data via the forward encoding process. Following training, the data is decoded iteratively from the latent space through the reverse generation process until the target data is attained.
	\subsection{Forward Encoding Training Process:}
	The forward encoding process gradually encodes samples from the real space to the Tth latent space\cite{pinaya2022brain}, ultimately transforming them into isotropic Gaussian noise\cite{sohl2015deep}. Given complete seismic data  $ x_0 \sim q(x_0)$ the forward encoding process encodes it into Gaussian noise $ x_T \sim \mathcal{N}(0,\mathbf{I})$ over T time steps\cite{song2019generative}. This process forms a Markov chain\cite{kingma2021variational} determined using a predefined variance table. The encoding process from step t-1 to t can be defined as:
	\begin{equation}
		 q(x_t \mid x_{t-1})=\mathcal{N}(\sqrt{1-\beta_t}x_{t-1},\beta_t\mathbf{I})
	\end{equation}
	Where $\beta_t$ is obtained from the existing variance table, typically set to linearly increase from 0.0001 to 0.002. 
		
		According to Ho et al.\cite{ho2020denoising}, Equation (1) can be further derived to obtain the encoding formula (2) from step 0 to step t:
	\begin{equation}
		q(x_t \mid x_{0})=\mathcal{N}(\sqrt{\overline{\alpha}_t}x_{0},(1-\overline{\alpha}_t)\mathbf{I})
	\end{equation}
	where $\alpha_t =1-\beta_t, \overline{\alpha}_t =\prod_{s=0}^t \alpha_s $.
	
	The diffusion model employs neural networks to model predictions and reverse this encoding process, aiming to obtain the probability distribution $p_\theta(x_{t-1}\mid x_{t})$ of the data at step t-1, as shown in Equation (3), where the mean $\mu_\theta(x_t,t)$ and variance $\beta_\theta(x_t,t)$ are obtained from neural networks.
	\begin{equation}
		p_\theta(x_{t-1} \mid x_{t})=\mathcal{N}(\mu_\theta(x_t,t),\beta_\theta(x_t,t))
	\end{equation}
	By minimizing the variational lower bound\cite{kingma2021variational} of the negative log-likelihood, we can obtain the loss function of the network:
	\begin{equation}
		 \begin{aligned}
				L_{vlb}&= \mathbb{E}_q[\underbrace{ D_{KL}(q(x_T \mid x_0) \parallel p(x_T)) }_{L_T} \\& +
				\sum_{t>1}\underbrace{ D_{KL}(q(x_t \mid x_{t-1},x_0) \parallel p_\theta(x_{t-1}\mid x_t)) }_{L_{t-1}}\\&-\underbrace{ \log p_\theta(x_0 \mid x_1) }_{L_0}]
		\end{aligned}
	\end{equation}
	Where $D_{KL}$ is the KL divergence, which describes the loss between the encoding distribution q and the network generated decoding distribution p.
		
		According to Ho et al.\cite{ho2020denoising}, during single step training, only the $L_{t-1}$ loss needs to be computed. Hence, a simplified loss function can be derived as shown in Equation (5), where $\epsilon_t$ is  is sampled from a Gaussian distribution and $\epsilon_\theta(x_t,t)$
		is predicted by network.
	\begin{equation}
		 L_{simple}=E_{t,x_0,\epsilon_t}[\lVert \epsilon_t-\epsilon_\theta(x_t,t) \rVert^2] 
	\end{equation}
	\begin{figure}[!t]
		\centering
		\includegraphics[width=2.1in]{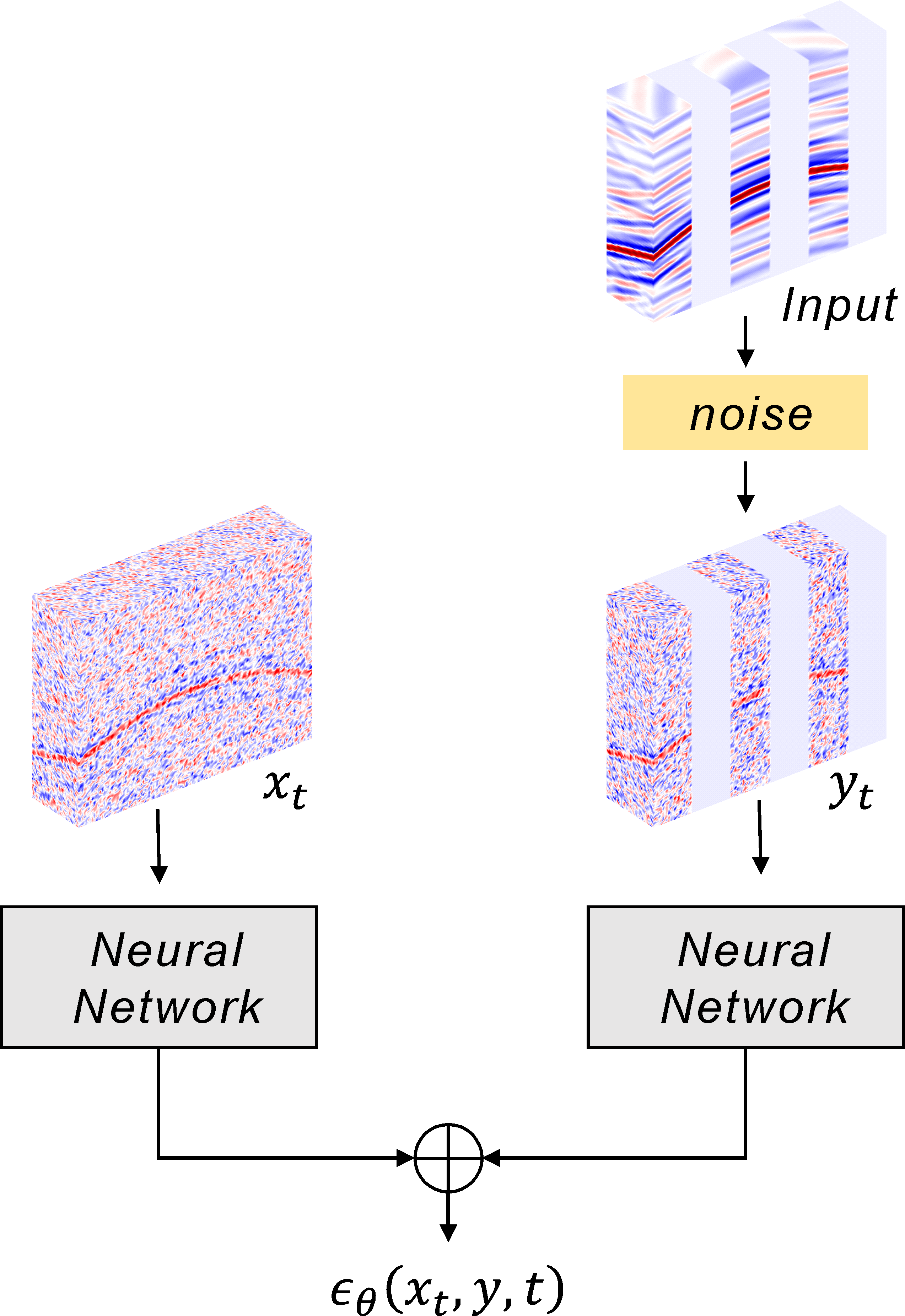}
		\caption{The overall framework of the noise matching network is as follows. $\epsilon_\theta(x_t,y,t)$ is composed of the sum of the output obtained from inputting $x_t$ into the network (on the left) and the output obtained from inputting $y_t$ into the network (on the right). $y_t$ represents known data obtained by adding noise through the forward encoding process.}
		\label{arcitect}
	\end{figure}
	\subsection{Reverse sampling generation process:}
	After training, the diffusion model starts decoding the data from the latent space iteratively until it reaches the target data, as shown in Fig. \ref{condition}. The decoding process from step t to step t-1 is as follows:
	\begin{equation}
		 p_\theta(x_{t-1} \mid x_{t})=\mathcal{N}(\frac {1}{\sqrt{\alpha_t}}(x_t - \frac {\beta_t}{\sqrt{1-\overline{\alpha}_t}}\epsilon_\theta(x_t,t)),\beta_\theta(x_t,t))
	\end{equation}
	Iterative sampling continues until obtaining the pure target data $x_{0}$
	\begin{figure*}[!t]
		\centering
		\includegraphics[width=0.8\textwidth]{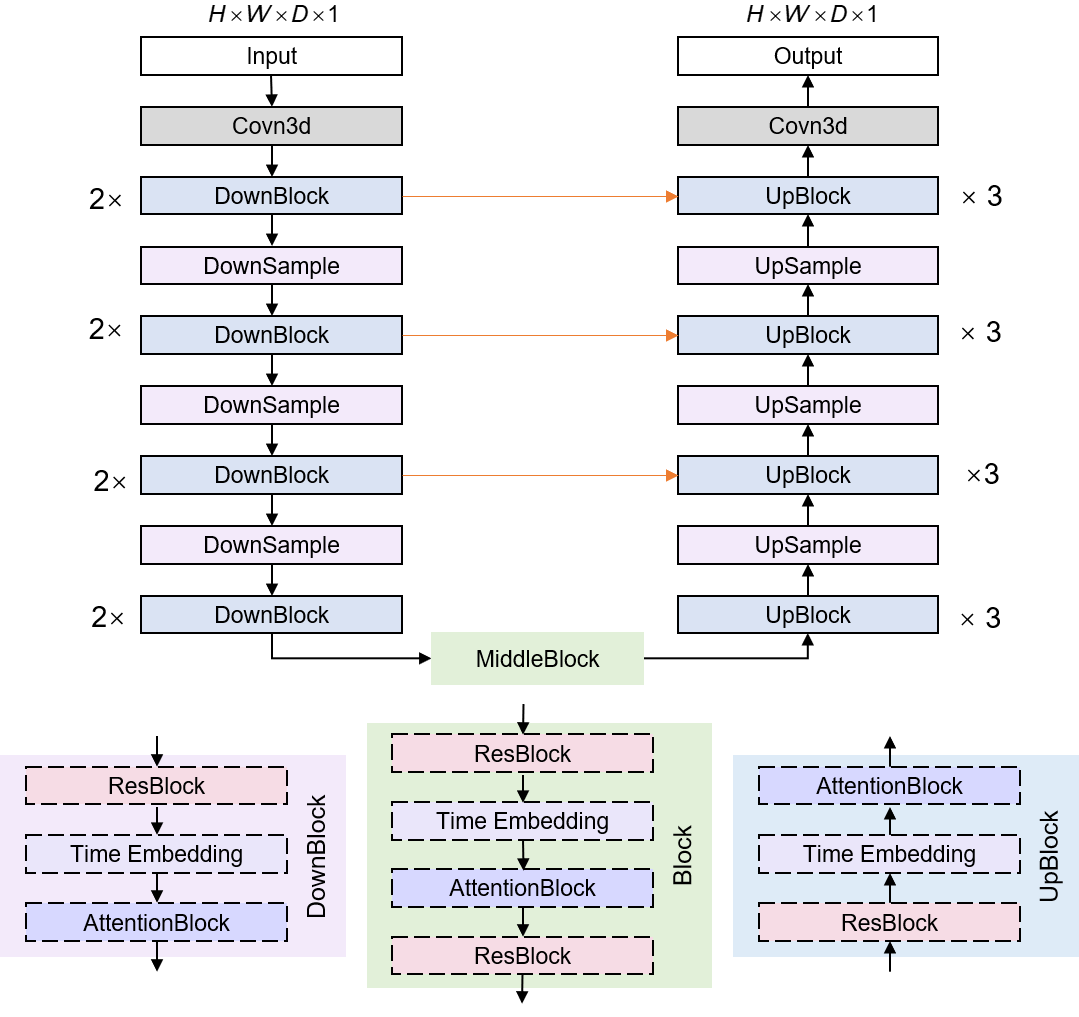}
		\caption{Backbone network architecture. On the left is the overall network architecture. The seismic data input undergoes a three-dimensional convolution, followed by downsampling through DownBlock and DownSample. After processing by MiddleBlock, it undergoes upsampling through UpBlock and UpSample, then passes through a three-dimensional convolution to obtain the output. On the right, from top to bottom, are the structures of DownBlock, MiddleBlock, and UpBlock.}
		\label{backbone}
	\end{figure*}
	\section{Method}
	\subsection{Constraint correction:}
	The diffusion model functions as a purely unconditional generative model. Following training, it can sample and generate data possessing a probability distribution akin to the training set. However, the entire sampling process remains unconstrained, potentially yielding diverse outcomes when sampling commences from distinct latent variables, albeit some may exhibit high consistency with the data intended for reconstruction. Hence, we aim to incorporate supervised constraints into the diffusion model. These constraints endeavor to align $x_{t-1}$, sampled at step $t$, with the (t-1)th latent space vector encoded by the ground truth. After undergoing T constraints, even when starting from different latent variables, the final sampling results will approximate the true values. The forward encoding process and the reverse sampling generation process with constraints are illustrated in Fig. \ref{condition}.
	
	In the forward encoding training process, after introducing self-supervised constraints, Equation (3) is rewritten as:
	\begin{equation}
		 p_\theta(x_{t-1} \mid x_{t})=\mathcal{N}(\mu_\theta(x_t,y,t),\beta_\theta(x_t,y,t))
	\end{equation}
	The value of $y$ is the self-supervised constraint we introduce into the constrained diffusion model.
	
	After introducing self-supervised constraints, the loss function derived from the variational lower bound considerations is rewritten as:
	\begin{equation}
		\begin{aligned}
			L_{vlb}&= \mathbb{E}_q[\underbrace{ D_{KL}(q(x_T \mid x_0) \parallel p(x_T)) }_{L_T} \\& + \sum_{t>1}\underbrace{ D_{KL}(q(x_t \mid x_{t-1,x_0}) \parallel p_\theta(x_{t-1}\mid x_t,y)) }\\&_{L_{t-1}}-\underbrace{ \log p_\theta(x_0 \mid x_1,y) }_{L_0}] 
		\end{aligned}
	\end{equation}
	The simplified loss function derived from $L_{t-1}$ is rewritten as:
	\begin{equation}
		L_{simple}=E_{t,x_0,\epsilon_t}[\lVert \epsilon_t-\epsilon_\theta(x_t,y,t) \rVert^2] 
	\end{equation}
	The network changes from predicting $\epsilon_\theta(x_t, t)$ to predicting $\epsilon_\theta(x_t, y, t)$. In the reverse sampling generation process, after introducing self-supervised constraints, the sampling at step t is:
	\begin{equation}
		p_\theta(x_{t-1} \mid x_{t})=\mathcal{N}(\frac {1}{\sqrt{\alpha_t}}(x_t - \frac {\beta_t}{\sqrt{1-\overline{\alpha}_t}}\epsilon_\theta(x_t,y,t)),\beta_\theta(x_t,y,t))
	\end{equation}
	
	\begin{figure*}[!t]
		\centering
		\includegraphics[width=0.8\textwidth]{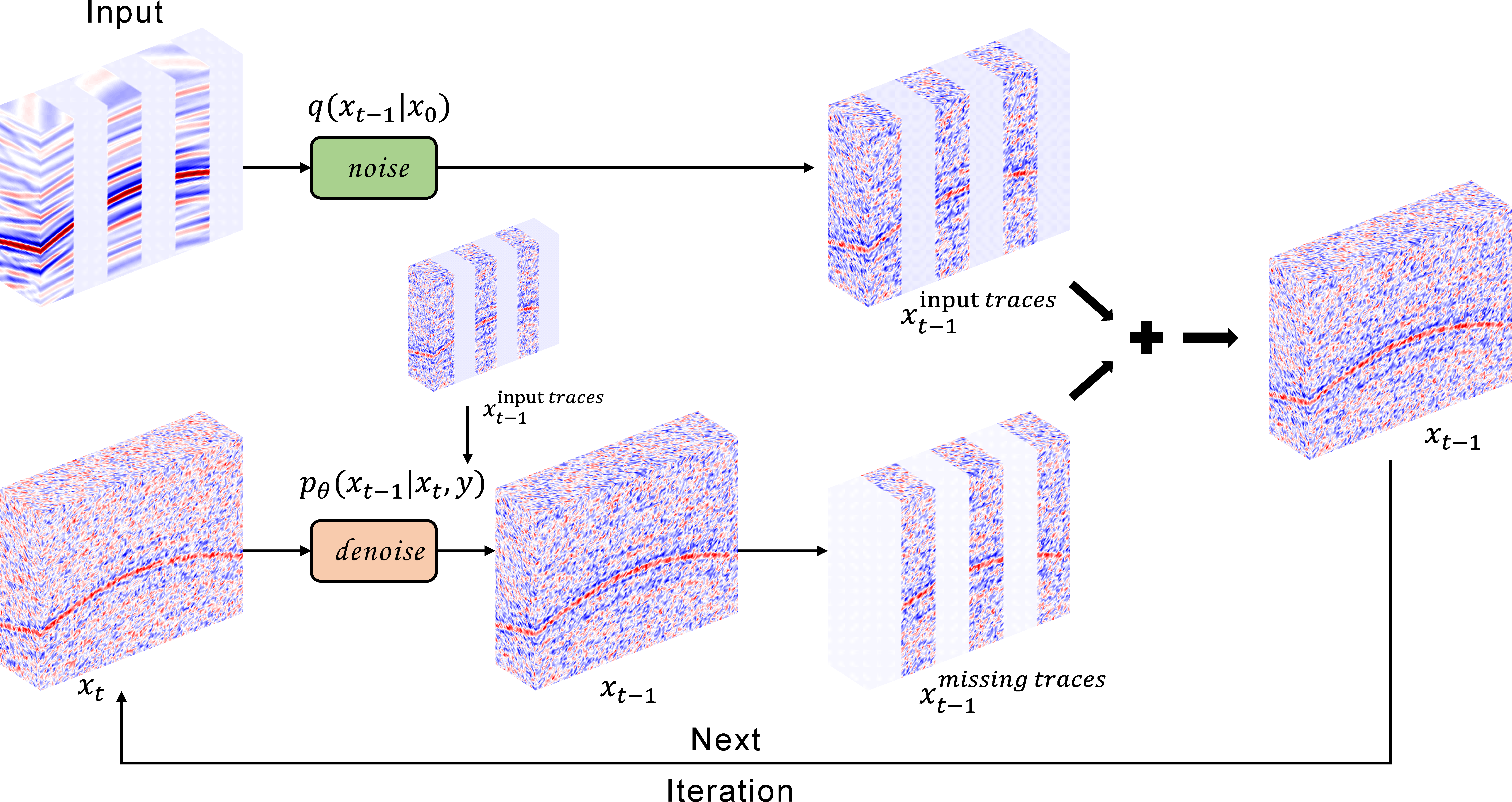}
		\caption{Guided sampling process. Guided sampling modifies the standard denoising process in order to condition on the given seismic data content. In each step, we sample the known region (top) from the input and the reconstruction part from the diffusion model output described in the Constraint Correction subsection(bottom).}
		\label{guided}
	\end{figure*}
	
	\subsection{3D neural network architecture:}
	In the diffusion model, neural networks are employed to estimate the noise scale, denoted as $\epsilon_\theta(x_t,y,t)$ in Equation (10). The precision in estimating $\epsilon_\theta(x_t,y,t)$ corresponding to the current step t directly impacts the quality of the generated data. Hence, extending the diffusion model from 2D to 3D preserves its original forward encoding training and reverse sampling generation principles. Consequently, its forward encoding process remains a fixed Markov chain, as depicted in Fig. \ref{condition}. Although the training objective still adheres to Equation (9), adjustments in the neural network architecture for generating $\epsilon_\theta(x_t,y,t)$ are necessary to render it suitable for 3D tasks.
	
	To tailor the diffusion model for 3D generation tasks, we introduce a 3D architecture with conditional constraints. The comprehensive architecture is outlined in Fig. \ref{arcitect}
	
	The network's prediction of $\epsilon_\theta(x_t,y,t)$ comprises two components: one entails the conditional features derived from the backbone network's output. The input to the backbone network, $y_t$, is obtained through the encoding of known traces. The other component involves input $x_t$ into the backbone network, which yields encoded features. Subsequently, these two components are fused to derive $\epsilon_\theta(x_t,y,t)$. The feature fused  $\epsilon_\theta(x_t,y,t)$ is then constrained by the input known data.
	
	The overall structure of the backbone network is shown in Fig. \ref{backbone}, which is a 3D UNet with attention and time embedding. The encoding part of the UNet consists of DownBlocks and DownSamples, where DownBlock deepens the channel without changing the size of the feature map. The decoding part of the UNet comprises UpBlocks and UpSamples, ensuring that the output and input maintain the same size dimensions. Each DownBlock and UpBlock performs time embedding to embed the information of the current step $t$, allowing the network to learn the information of $t$ to the fullest extent and enabling more precise prediction of the encoded features and conditional features by the network.
	\begin{figure}[!t]
		\centering
		\includegraphics[width=2in]{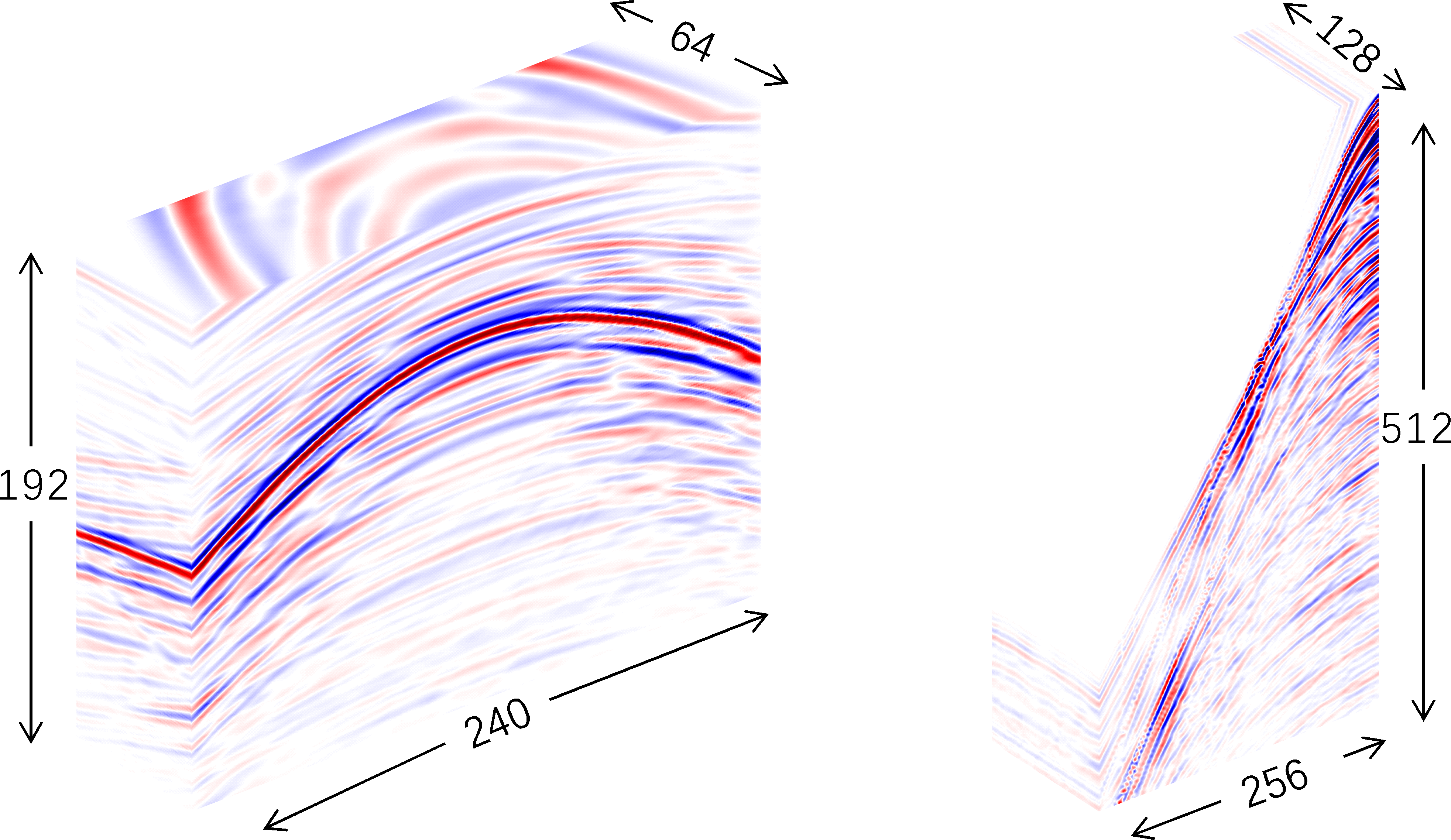}
		\caption{Dataset, left is SEG C3, Right is Mobil Avo Viking Graben Line 12.}
		\label{dataset}
	\end{figure}
	\begin{figure*}[!t]
		\centering
		\includegraphics[width=0.8\textwidth]{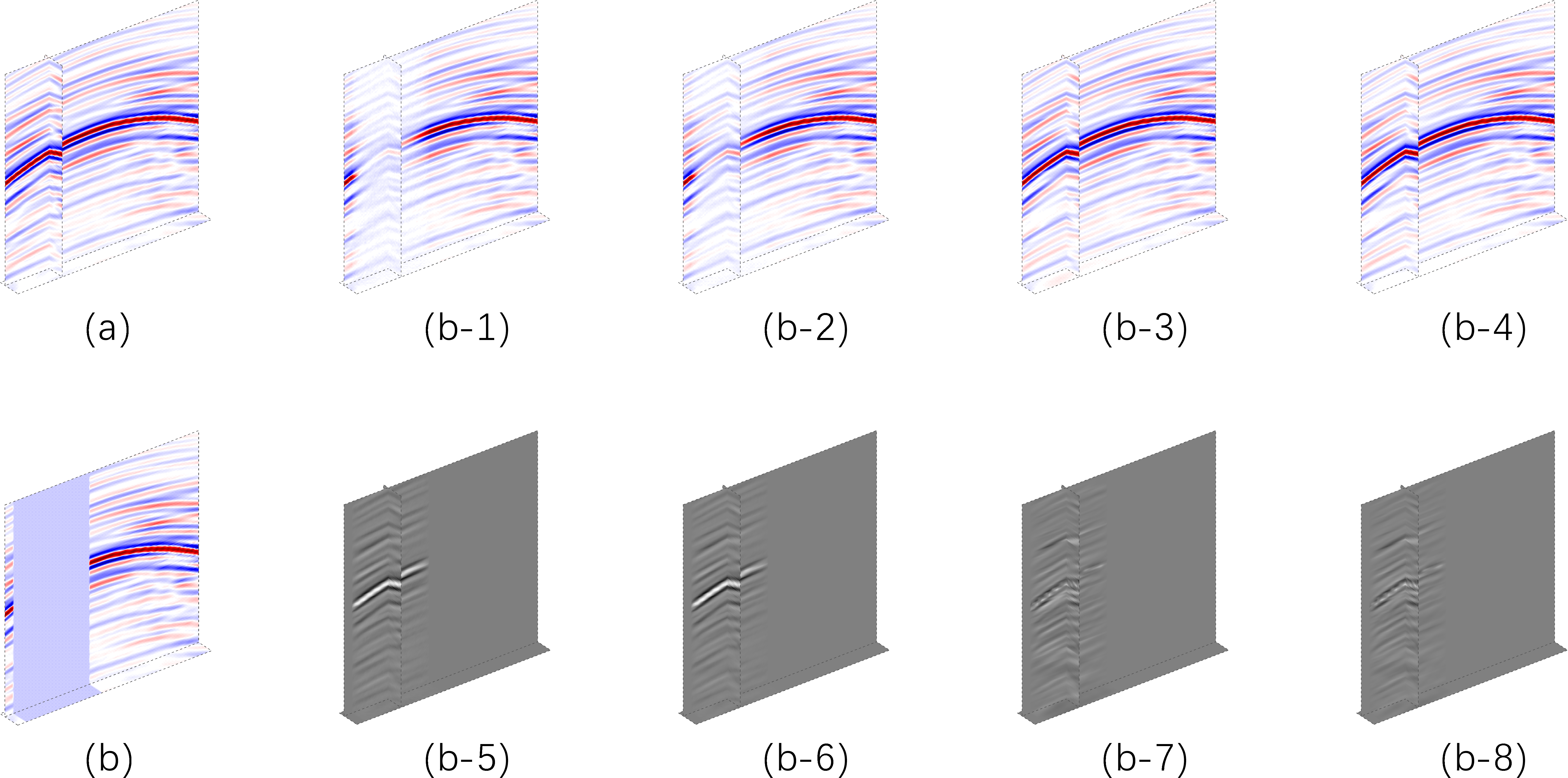}
		\caption{(a) Ground truth, (b) 50\% 50 continuous missing traces, (b-1)-(b-4) are the reconstruction results with u-values set to 1, 5, 10, and 15, respectively, and (b-5)-(b-8) are the residuals of each reconstruction result relative to Ground truth, respectively.}
		\label{xiaor}
	\end{figure*}
	\subsection{Guided sampling}
	The diffusion model described in Section \ref{ddpm} is an unconditional generative model. It iteratively decodes samples from the $T$th  latent space until the target data is obtained. This sampling process does not accept any prior information. Therefore, directly using the diffusion model for seismic data reconstruction leads to conflicts between the probability distribution of the generated data and that of the data to be reconstructed. Consequently, the reconstruction results struggle to maintain high consistency with the data to be reconstructed, which is undesirable for data reconstruction tasks. The guided sampling approach we propose effectively addresses this issue.
	
	When the diffusion model samples $x_{t-1}$ from $x_t$, if $x_t$ contains prior information about known data, then the generated $x_{t-1}$ naturally includes the prior information of the known data\cite{lugmayr2022repaint}. As a result, the iteratively sampled $x_0$ will exhibit better consistency with the known data. In guided sampling, we represent seismic data as $x$, where the known portion is denoted as $(1-m) \odot x$ and the portion to be reconstructed is denoted as $m \odot x$. Then make the following improvement in sampling $x_{t-1}$ from $x_t$:
	\begin{subequations}
		\begin{equation}
			x_{t-1}^{known} \sim \mathcal{N}(\sqrt{\overline{\alpha}_{t-1}}x_0,(1-\overline{\alpha}_{t-1})\mathbf{I})
		\end{equation}
		\begin{equation}
			x_{t-1}^{unknown} \sim \mathcal{N}(\mu_\theta(x_t,y,t),\beta_\theta(x_t,y,t)\mathbf{I})
		\end{equation}
		\begin{equation}
			x_{t-1}=(1-m)\odot x_{t-1}^{known}+m\odot x_{t-1}^{unknown}
		\end{equation}
	\end{subequations}
	Thus, $x_{t-1}$ consists of two parts: one part is encoded from the known data to the ${t-1}$ latent space through the forward encoding process, and the other part is sampled from $x_t$ during the sampling process. At this point, $x_{t-1}$ contains the prior information of the known data. However, $x_{t-1}$ still lacks consistency with the known data, and continuing to sample $x_{t-2}$ will not resolve this conflict. However, we can encode $x_{t-1}$ back into $x_t$ and obtain $x_{t-1}$ again using the above method. In this case, $x_{t-1}$ will have higher consistency. We repeat this process, and the resulting $x_{t-1}$ will no longer have consistency conflicts. The overall process is illustrated in Fig. \ref{guided}, and specific steps are listed in Algorithm 1.
	
	\begin{algorithm}[H]
		\caption{Seismic data reconstruction algorithm.}\label{alg:cap}
		\begin{algorithmic}
			\STATE
			\STATE \textbf{ for } {$t=T,\ldots,1$}
			\STATE \hspace{0.5cm} \textbf{ for } {$u=1,\ldots,U$}
			\STATE \hspace{1.0cm}$\epsilon\sim \mathcal{N}(0,I)$ if $t>1$,else $\epsilon=0$
			\STATE \hspace{1.0cm}$x_{t-1}^{known} =\sqrt{\overline{\alpha}_{t}}x_0+\epsilon\sqrt{(1-\overline{\alpha}_{t})})$
			\STATE \hspace{1.0cm}$z\sim \mathcal{N}(0,I)$ if $t>1$,else $z=0$
			\STATE \hspace{1.0cm}$x_{t-1}^{unknown} =\frac {1}{\sqrt{\alpha_t}}(x_t - \frac {\beta_t}{\sqrt{1-\overline{\alpha}_t}}\epsilon_\theta(x_t,y,t))+z\sigma_t$
			\STATE \hspace{1.0cm}$x_{t-1}=(1-m)\odot x_{t-1}^{known}+m\odot x_{t-1}^{unknown}$
			\STATE \hspace{1.0cm}\textbf{ if } {$u<U$ and $t>1$}\textbf{ then }
			\STATE \hspace{1.5cm}$x_t\sim \mathcal{N}(\sqrt{1-\beta_{t-1}}x_{t-1},\beta_{t-1}\mathbf{I})$
			\STATE \hspace{1.0cm} \textbf{ end if }
			\STATE \hspace{0.5cm} \textbf{ end for }
			\STATE \textbf{ end for }
			\STATE \textbf{ return }{$x_0$}
		\end{algorithmic}
	\end{algorithm}
	\section{Experiments}
	\subsection{Evaluation Metrics:}
	To quantitatively evaluate the quality of the reconstruction results, we selected three commonly used metrics, namely MSE, SNR, and SSIM. MSE measures the error between the reconstruction result and the ground truth, calculated using the following formula:
	\begin{equation}
		MSE=\frac {1}{n}\sum_{i=1}^{n}(x_{i}^{r}-x_{i}^{t})^2
	\end{equation}
	Where $x_{i}^{r}$ represents the data reconstructed by the network, $x_{i}^{t}$ represents the ground truth, and the closer the MSE value is to 0, the closer the reconstruction result is to the ground truth. SNR measures the quality of the reconstruction result, calculated by the following formula:
	\begin{equation}
		SNR=10log_{10}\frac{\parallel x_r \parallel_F^2 }{\parallel x_t-x_r \parallel_F^2} 
	\end{equation}
	Where $x_r$ represents the data reconstructed by the network, $x_t$ represents the ground truth, $\parallel \parallel$ denotes the Frobenius norm, and the higher the SNR value, the higher the quality of the reconstruction result. SSIM measures the structural similarity of the reconstruction result, calculated by the following formula:
	\begin{equation}
		SSIM=\frac{(2\mu_r\mu_t+c_1)(2\sigma _{rt}+c_2)}{(\mu_r^2\mu_t^2+c_1)(\sigma _r^2+\sigma _t^2+c_2)} 
	\end{equation}
	Where $\mu_r$ is the mean of the reconstructed result, $\mu_t$ is the mean of the ground truth, $2\sigma _{rt}$ is the covariance between the reconstructed result and the ground truth, $\sigma _r$ is the variance of the reconstructed result, $\sigma_t$ is the variance of the ground truth, and $c_1$ and $c_2$ are two constants introduced to avoid numerical instability. The SSIM value closer to 1 indicates that the reconstruction result is closer to the ground truth.
	\begin{figure*}[!t]
		\centering
		\includegraphics[width=0.8\textwidth]{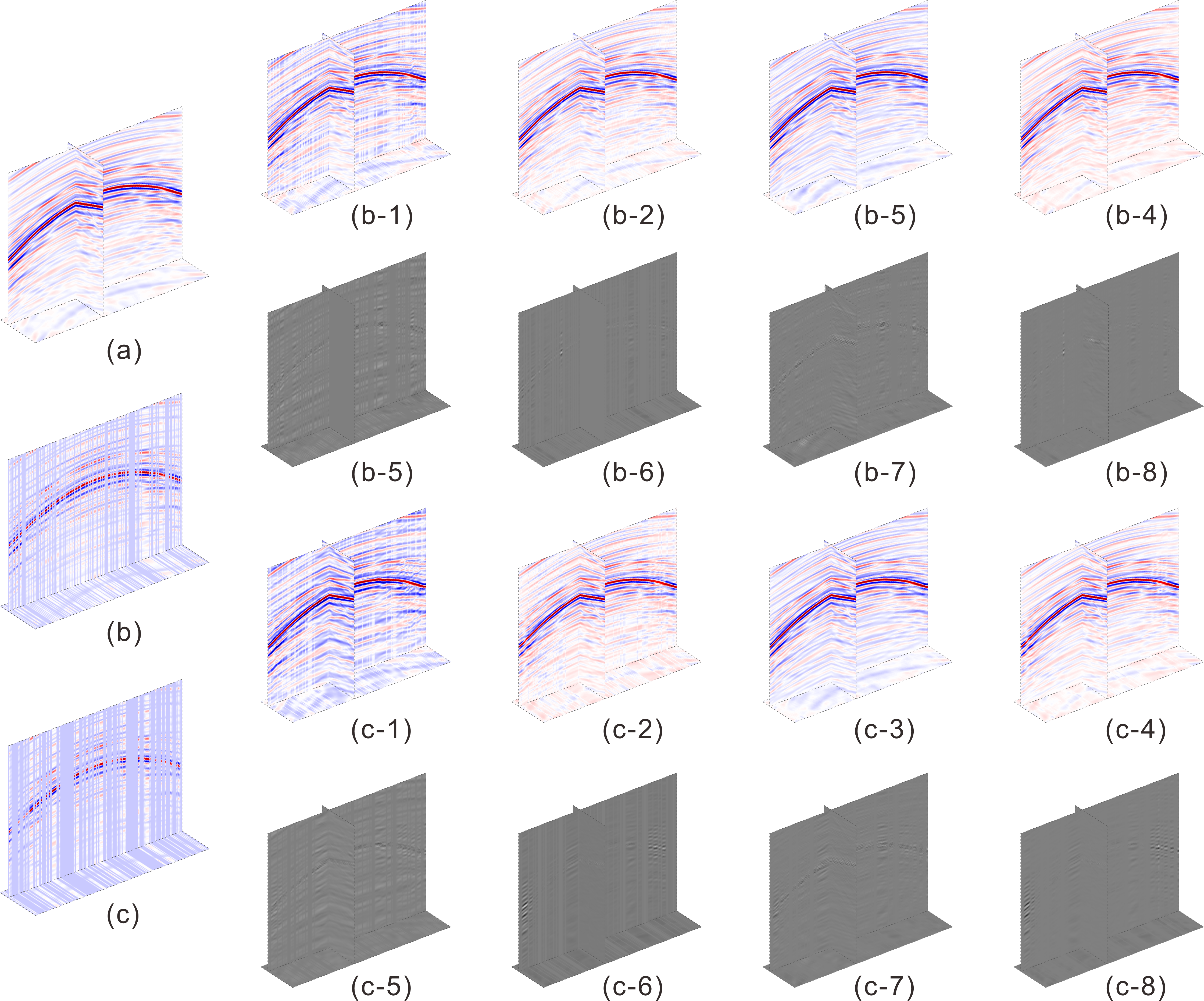}
		\caption{(a) Ground truth,(b) 50\% random missing, (b-1) to (b-4) are the reconstruction results of Unet, conditional DPM, MDA GAN, and SeisFusion, respectively. (b-5) to (b-8) are the residuals of each reconstruction result relative to the ground truth.(c) 80\% random missing, (c-1) to (c-4) are the reconstruction results of Unet, conditional DPM, MDA GAN, and SeisFusion, respectively. (c-5) to (c-8) are the residuals of each reconstruction result relative to the ground truth.}
		\label{synthetic}
	\end{figure*}
	\subsection{Train:}
	Our experiments were conducted on two datasets as shown in Fig. \ref{dataset}, the publicly available synthetic dataset SEG C3 and the field dataset Mobil Avo Viking Graben Line 12. We divided the datasets into three parts: 3/5 for training, 1/5 for validation, and 1/5 for testing. The patch size was set to 16x32x128. Randomly missing training samples were used as constraint inputs. We set T, the number of iterations in the forward process, to 1000. Adam was chosen as the optimization algorithm with a learning rate of 1e-5. The models were trained for 500,000 steps on two RTX3090Ti GPUs, and the model with the lowest loss on the validation set was selected as the final model for testing.
	
	\subsection{Ablation:}
	In order to reasonably determine the values for the parameter $U$  in Algorithm 1, we first conducted ablation studies. The ablation studies were conducted on a patch of size 16×128×128 from the SEG C3 dataset, where 50\% of contiguous traces were set to 0 to represent missing traces. Different values of $U$ were set for reconstruction, and the results are shown in Fig. \ref{xiaor}.
	
	When the value of $U$ is set to 1, no iterative guided sampling is performed. The reconstruction result is generated by a constrained diffusion model without guided sampling constraints. It can be observed that the generated results at this point no longer exhibit randomness and have been constrained to a reasonable range, although the generation details still need improvement. As the value of $U$ increases to 5, the reconstruction details noticeably improve. Further increasing $U$ to 10, the reconstruction quality reaches a bottleneck, showing higher consistency with the ground truth. We computed evaluation metrics for four values of $U$  as shown in Table~\ref{tab:Ablation} to quantitatively assess the influence of $U$ on the reconstruction results. It can be seen that as $U$  increases, the reconstruction performance gradually improves, reaching a bottleneck at $U=10$. Increasing $U$ further does not significantly enhance the reconstruction quality. Therefore, setting $U$ to 10 is a reasonable choice.
	\begin{table}[!t]
		\caption{Ablation study.}\label{tab:Ablation}
		\centering
		\linespread{1.25} \selectfont
		\begin{tabular}{cccc}
			\hline
			U  & MSE       & SNR     & SSIM   \\ \hline
			1  & 1.7282e-3 & 27.6239 & 0.9196 \\
			5  & 1.1313e-3 & 29.4639 & 0.9510 \\
			10 & 1.6460e-4 & 37.8356 & 0.9875 \\
			15 & 1.7072e-4 & 37.6770 & 0.9880 \\ \hline
		\end{tabular}
	\end{table}
	\begin{figure*}[!t]
		\centering
		\includegraphics[width=0.8\textwidth]{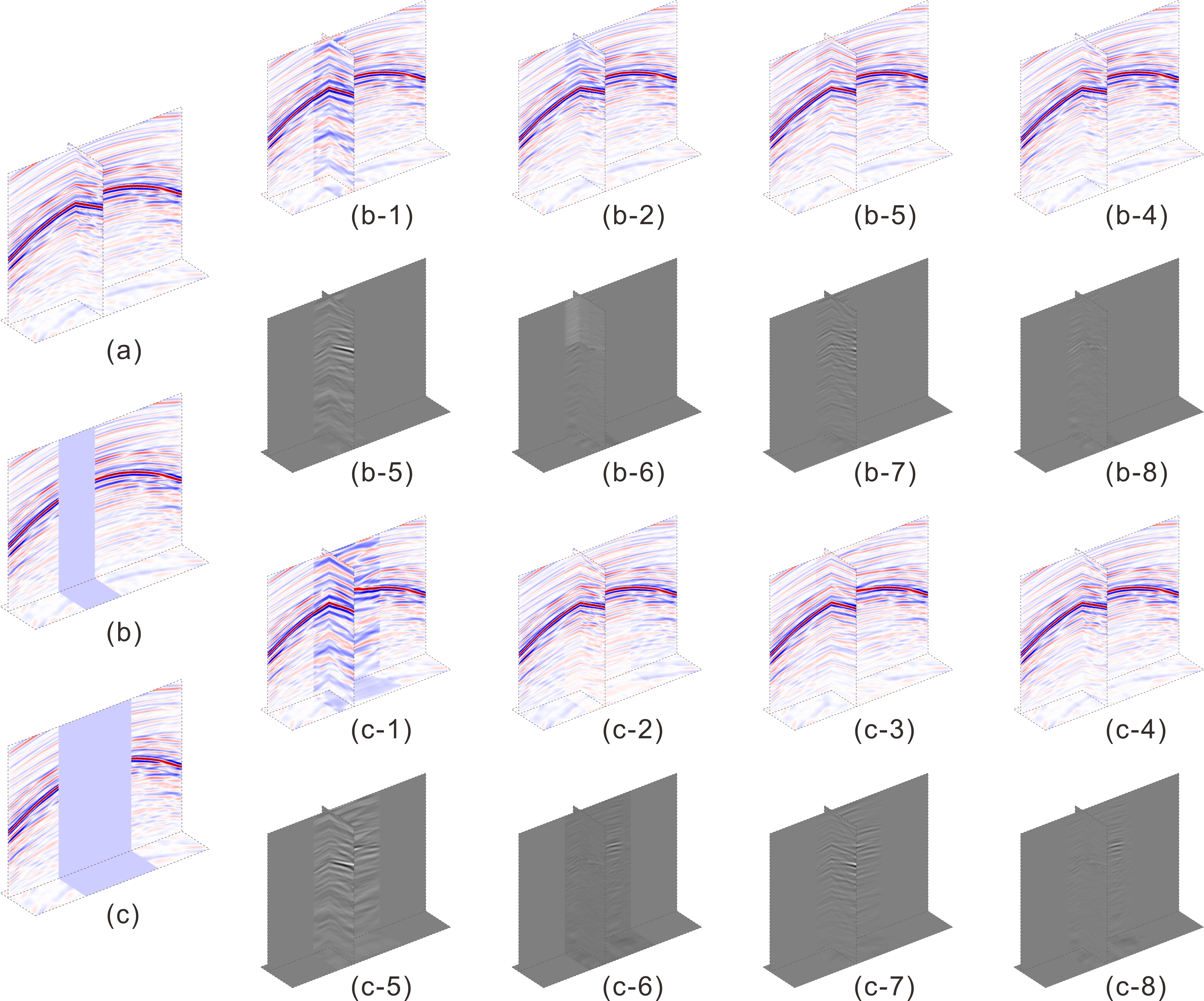}
		\caption{(a) Ground truth,(b) 50 continuous missing traces, (b-1) to (b-4) are the reconstruction results of Unet, conditional DPM, MDA GAN, and SeisFusion, respectively. (b-5) to (b-8) are the residuals of each reconstruction result relative to the ground truth.(c) 100 continuous missing traces, (c-1) to (c-4) are the reconstruction results of Unet, conditional DPM, MDA GAN, and SeisFusion, respectively. (c-5) to (c-8) are the residuals of each reconstruction result relative to the ground truth.}
		\label{synthetic_b}
	\end{figure*}
		\begin{figure*}[!t]
		\centering
		\includegraphics[width=0.8\textwidth]{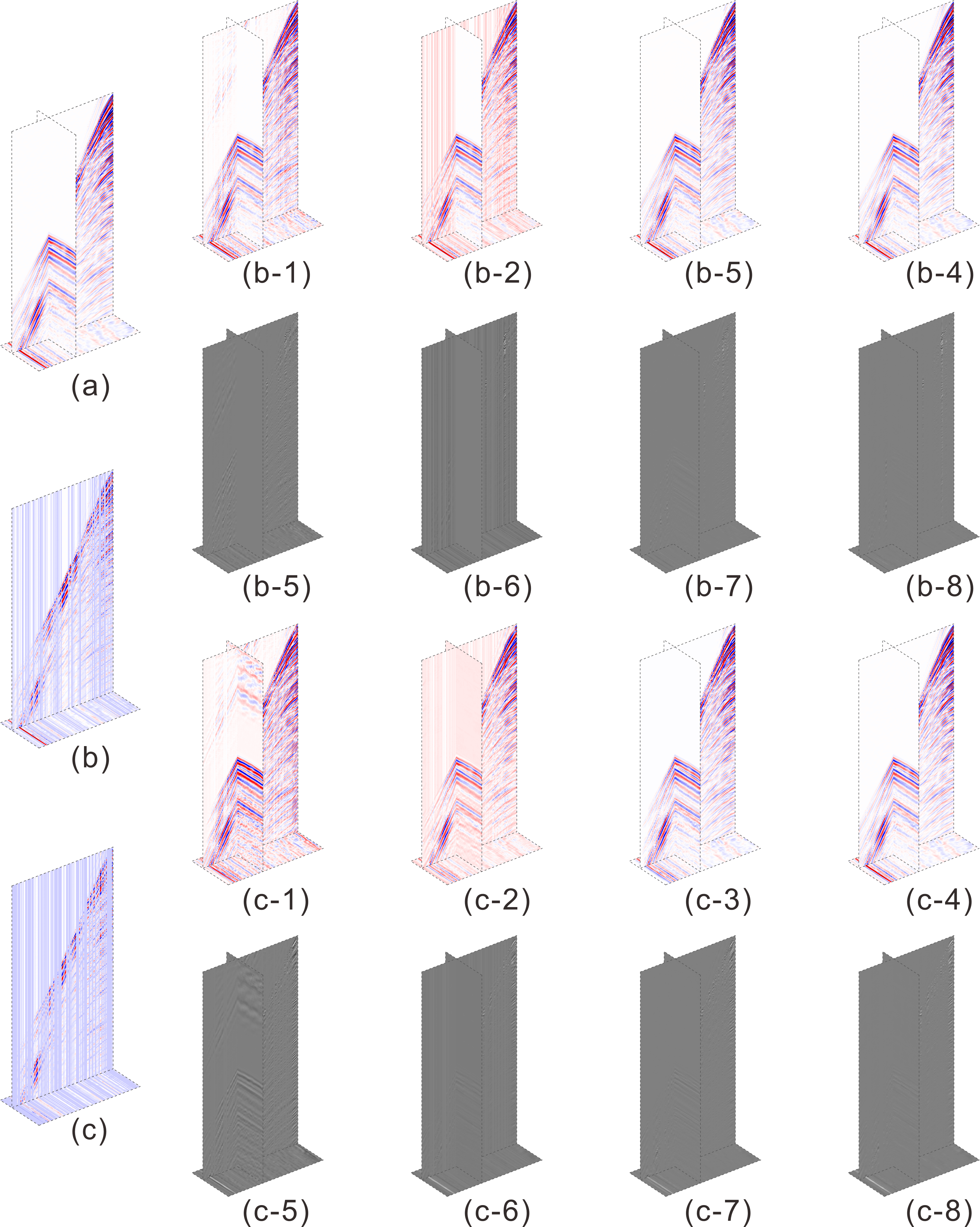}
		\caption{(a) Ground truth,(b) 50\% random missing, (b-1) to (b-4) are the reconstruction results of Unet, conditional DPM, MDA GAN, and SeisFusion, respectively. (b-5) to (b-8) are the residuals of each reconstruction result relative to the ground truth.(c) 80\% random missing, (c-1) to (c-4) are the reconstruction results of Unet, conditional DPM, MDA GAN, and SeisFusion, respectively. (c-5) to (c-8) are the residuals of each reconstruction result relative to the ground truth.}
		\label{filed}
	\end{figure*}
	
	\subsection{Data Reconstruction:}
	During the sampling generation process, in Algorithm 1, T is set to 250 based on the state-of-the-art diffusion model, and $U$ is set to 10. To validate the effectiveness of our proposed method, we selected three other models for comparative testing: Unet, conditional DPM, and MDA GAN. Unet is the most commonly used model, which has been proven to have excellent performance in seismic data reconstruction. We redesigned the UNet following the methodology of Anet\cite{yu2021attention} and trained it using the same dataset. However, since the related studies did not release their code, we could not ascertain the exact hyperparameters. We trained the model to its optimal state according to the parameters set in this paper, using Adam as the optimization algorithm with a learning rate of 1e-4. The training was conducted for 300 epochs on two RTX3090Ti GPUs. The conditional DPM\cite{liu2024generative} is two-dimensional and cannot be used for three-dimensional reconstruction, so we made some modifications by changing their 2D architecture to a 3D architecture, enabling it to perform 3D reconstruction. For MDA GAN\cite{dou2023mda}, the authors provided the source code and weight files, stating that they did not need to be retrained. Therefore, we directly used the code and weights provided by the authors for reconstruction without any modifications.
	\subsubsection{SEG C3}
	To validate the effectiveness of our method, we first conducted tests on the synthetic dataset SEG C3. To fully evaluate the reconstruction performance of the proposed method under different missing scenarios, we divided the tests into two parts: random discrete missing and continuous missing.
		\begin{figure*}[!t]
		\centering
		\includegraphics[width=0.8\textwidth]{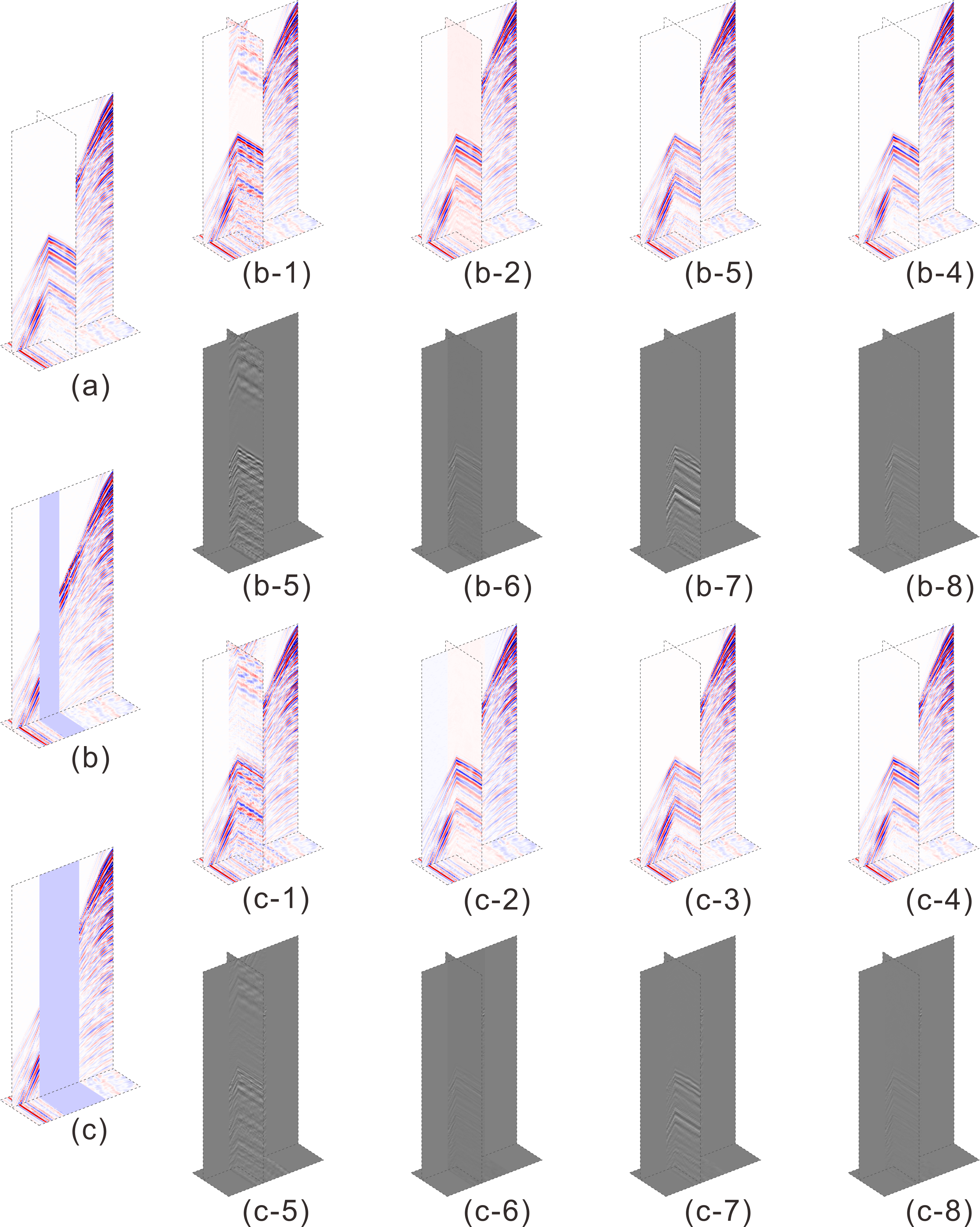}
		\caption{(a) Ground truth,(b) 50 continuous missing traces, (b-1) to (b-4) are the reconstruction results of Unet, conditional DPM, MDA GAN, and SeisFusion, respectively. (b-5) to (b-8) are the residuals of each reconstruction result relative to the ground truth.(c) 100 continuous missing traces, (c-1) to (c-4) are the reconstruction results of Unet, conditional DPM, MDA GAN, and SeisFusion, respectively. (c-5) to (c-8) are the residuals of each reconstruction result relative to the ground truth.}
		\label{filed_b}
	\end{figure*}
	Random discrete missing: To simulate possible complex situations, we set two different missing rates, 50\% and 80\%, by setting traces to 0 to represent missing data. Fig. \ref{synthetic} shows the reconstruction results of different methods under these two missing rates. It can be observed that the reconstruction results of Unet already show low consistency at a 50\% missing rate, and the discontinuity becomes more pronounced when the missing rate increases to 80\%. Residual plots indicate a significant deviation between its reconstruction results and ground truth. On the other hand, when reconstructing three-dimensional data, conditional DPM experiences a slight performance decrease due to the increased complexity of the data. Both MDA GAN and our method perform well in reconstructing data under both missing scenarios, with a slight advantage of our method over MDA GAN, as indicated by residual plots.
	
	To quantitatively evaluate the reconstruction performance of the four models, we computed four evaluation metrics, as shown in Table~\ref{tab:synthetic}. It can be seen that our method achieves the best results under different missing rates, with a significant lead in both MSE and SNR metrics.
	\begin{table}[!t]
		\caption{Comparison of four reconstruction networks under random missing on synthetic dataset.}\label{tab:synthetic}
		\centering
		\linespread{1.25} \selectfont
		\begin{tabular}{ccccc}
			\hline
			&                 & MSE       & SNR     & SSIM   \\ \hline
			\multirow{4}{*}{\begin{tabular}[c]{@{}c@{}}Random Missing\\ 50\% Traces\end{tabular}} & UNet            & 1.2044e-4 & 36.8362 & 0.9651 \\
			& Conditional DPM & 8.9571e-5 & 39.2495 & 0.9864 \\
			& MDA GAN         & 1.2719e-4 & 38.5153 & 0.9823 \\
			& SeisFusion        & \pmb{2.3166e-5} & \pmb{46.3513} & \pmb{0.9987} \\ \hline
			\multirow{4}{*}{\begin{tabular}[c]{@{}c@{}}Random Missing\\ 80\% Traces\end{tabular}} & UNet            & 8.2702e-4 & 32.2894 & 0.9451 \\
			& Conditional DPM & 4.8769e-4 & 33.1184 & 0.9791 \\
			& MDA GAN         & 2.9637e-4    & 35.2825 & 0.9789 \\
			& SeisFusion        & \pmb{2.2795e-4} & \pmb{36.4215} & \pmb{0.9881} \\ \hline
		\end{tabular}
	\end{table}

	Continuous missing: To evaluate the reconstruction performance of the proposed method under continuous missing, we set two scenarios of continuous missing, 50 continuous missing traces and 100 continuous missing traces. Traces were set to 0 to represent missing data. The reconstruction results of the four models are shown in Fig. \ref{synthetic_b}. It can be observed that UNet fails to reconstruct seismic data under continuous missing, showing poor reconstruction performance. When reconstructing such continuous missing data, conditional DPM exhibits a bias towards higher values compared to the original data. According to the residual maps, significant deviations from the ground truth can be observed in some areas. MDA GAN achieves relatively good reconstruction results, but deviations and inconsistencies still appear at the missing boundaries. Our method consistently provides the best reconstruction results under different missing scenarios.

	To quantitatively evaluate the reconstruction performance of the four models, we calculate four evaluation metrics under continuous missing, as shown in Table~\ref{tab:synthetic_b}. It can be seen that our method achieves the best results under different missing rates, with significant advantages in both MSE and SNR metrics. Especially under 100 continuous missing traces, our method also maintains a significant lead in the SSIM metric.
	\begin{table}[!t]
		\caption{Comparison of four reconstruction networks under continuous missing on synthetic dataset.}\label{tab:synthetic_b}
		\centering
		\linespread{1.25} \selectfont
		\begin{tabular}{ccccc}
			\hline
			&                 & MSE       & SNR     & SSIM   \\ \hline
			\multirow{4}{*}{\begin{tabular}[c]{@{}c@{}}50 continuous \\ missing traces\end{tabular}}  & UNet            & 1.7565e-3    & 27.5535 & 0.9328 \\
			& Conditional DPM & 5.1959e-4 & 32.1184 & 0.9613 \\
			& MDA GAN         & 3.5121e-4 & 34.5442 & 0.9767 \\
			& SeisFusion        & \pmb{1.9208e-5} & \pmb{47.1651} & \pmb{0.9988} \\ \hline
			\multirow{4}{*}{\begin{tabular}[c]{@{}c@{}}100 continuous \\ missing traces\end{tabular}} & UNet            & 4.0417e-3    & 23.9342  & 0.8414 \\
			& Conditional DPM & 8.7652e-4    & 25.4631 & 0.9084 \\
			& MDA GAN         & 2.0235e-4 & 26.9389 & 0.9027 \\
			& SeisFusion        & \pmb{6.2141e-5} & \pmb{42.0662} & \pmb{0.9965} \\ \hline
		\end{tabular}
	\end{table}
	\subsubsection{Mobil Avo Viking Graben Line 12}
	In order to further evaluate the performance of the proposed method and validate the effectiveness of the approach, we proceeded with testing on the Mobil Avo Viking Graben Line 12 field dataset. The testing was divided into two parts: random discrete missing and Continuous Missing.
	
	Random discrete missing: we set two different missing rates, 50\% and 80\%, where traces were set to 0 to represent the missing data. Fig. \ref{filed} displays the reconstruction results of different methods under these two missing rates. It can be observed that the reconstruction results of UNet and the conditional DPM already show low consistency at 50\% missing rate, and as the missing rate increases, the inconsistency becomes more apparent, as indicated by the residual plots showing significant deviations from the ground truth. Both MDA GAN and our method performed well in reconstructing under both missing scenarios, with our method slightly outperforming MDA GAN according to the residual plots.
	
	To quantitatively evaluate the reconstruction results of the four models, we calculated four evaluation metrics as shown in Table~\ref{tab:filed}. It can be seen that our method achieved the best performance under different missing rates, with a significant lead in both MSE and SNR evaluation 
	\begin{table}[!t]
		\caption{Comparison of four reconstruction networks under random missing on filed dataset.}\label{tab:filed}
		\centering
		\linespread{1.25} \selectfont
		\begin{tabular}{ccccc}
			\hline
			&                 & MSE       & SNR     & SSIM   \\ \hline
			\multirow{4}{*}{\begin{tabular}[c]{@{}c@{}}Random Missing\\ 50\% Traces\end{tabular}} & UNet            & 7.4965e-4 & 31.2513 & 0.9358 \\
			& Conditional DPM & 1.0068e-3 & 29.9702 & 0.9213 \\
			& MDA GAN         & 2.3697e-4 & 36.2529 & 0.9900 \\
			& SeisFusion        & \pmb{9.1223e-5} & \pmb{40.3989} & \pmb{0.9961} \\ \hline
			\multirow{4}{*}{\begin{tabular}[c]{@{}c@{}}Random Missing\\ 80\% Traces\end{tabular}} & UNet & 1.6030e-3 & 27.9505 & 0.8987 \\
			& Conditional DPM  & 3.6524e-3 & 26.5859 & 0.8947 \\
			& MDA GAN         & 5.6353e-4 & 32.4908 & 0.9772 \\
			& SeisFusion        & \pmb{3.0559e-4} & \pmb{35.1493} & \pmb{0.9891} \\ \hline
		\end{tabular}
	\end{table}
	
	Continuous missing: we similarly set two scenarios: continuous missing of 50 traces and continuous missing of 100 traces. Traces were set to 0 to represent the missing data, and the reconstruction results of the four models are shown in Fig. \ref{filed_b}. It can be observed that UNet fails to reconstruct seismic data under continuous missing, reconstructing undesired data in the low-amplitude part, and the conditional DPM reconstructed data show some inconsistency with high values overall. MDA GAN performs relatively well in reconstruction, but deviations and inconsistencies still appear at the missing boundaries. Our method provides optimal reconstruction results under different missing scenarios.
	
	To quantitatively evaluate the reconstruction results of the four models, we calculated four evaluation metrics under continuous missing, as shown in Table~\ref{tab:filed_b}. It can be seen that our method achieves the best performance under different missing rates, with a significant lead in both MSE and SNR evaluation metrics. Particularly, under 100-trace continuous missing, our method also maintains a substantial lead in the SSIM metric.
	\begin{table}[!t]
		\caption{Comparison of four reconstruction networks under continuous missing on filed dataset.}\label{tab:filed_b}
		\centering
		\linespread{1.25} \selectfont
		\begin{tabular}{ccccc}
			\hline
			&                 & MSE       & SNR     & SSIM   \\ \hline
			\multirow{4}{*}{\begin{tabular}[c]{@{}c@{}}50 continuous\\ missing traces\end{tabular}}  & UNet            & 4.6285e-4    & 33.3455 & 0.9617 \\
			& Conditional DPM & 3.0986e-4 & 34.5915 & 0.9649 \\
			& MDA GAN         & 2.9249e-4 & 35.3388 & 0.9774 \\
			& SeisFusion        & \pmb{6.4218e-5} & \pmb{41.9234} & \pmb{0.9945} \\ \hline
			\multirow{4}{*}{\begin{tabular}[c]{@{}c@{}}100 continuous\\ missing traces\end{tabular}} & UNet         & 1.3854e-3    & 28.5839  & 0.9015 \\
			& Conditional DPM & 1.9679e-3    & 29.0143 & 0.9263 \\
			& MDA GAN         & 1.1688e-3 & 29.3224 & 0.9390 \\
			& SeisFusion        & \pmb{3.0648e-4} & \pmb{35.1359} & \pmb{0.9848} \\ \hline
		\end{tabular}
	\end{table}
	
	The experiments on synthetic and filed datasets indicate that as the complexity of missing traces increases, the reconstruction performance experiences varying degrees of decline. This is largely because when the seismic data reconstruction task extends from 2D to 3D, the diversity of data exhibits exponential growth, whereas convolutional neural networks performing point-to-point reconstruction cannot achieve comprehensive distribution coverage. Therefore, with increasing complexity of missing traces, particularly in reconstructing continuous large missing segments, the performance degradation becomes more pronounced. In contrast, the diffusion model learns the probability distribution of the target data and can provide ideal characteristics of distribution coverage. Thus, our method maintains relatively good reconstruction accuracy when facing complex missing scenarios, especially continuous large missing segments.
		\begin{figure*}[!t]
		\centering
		\includegraphics[width=0.8\textwidth]{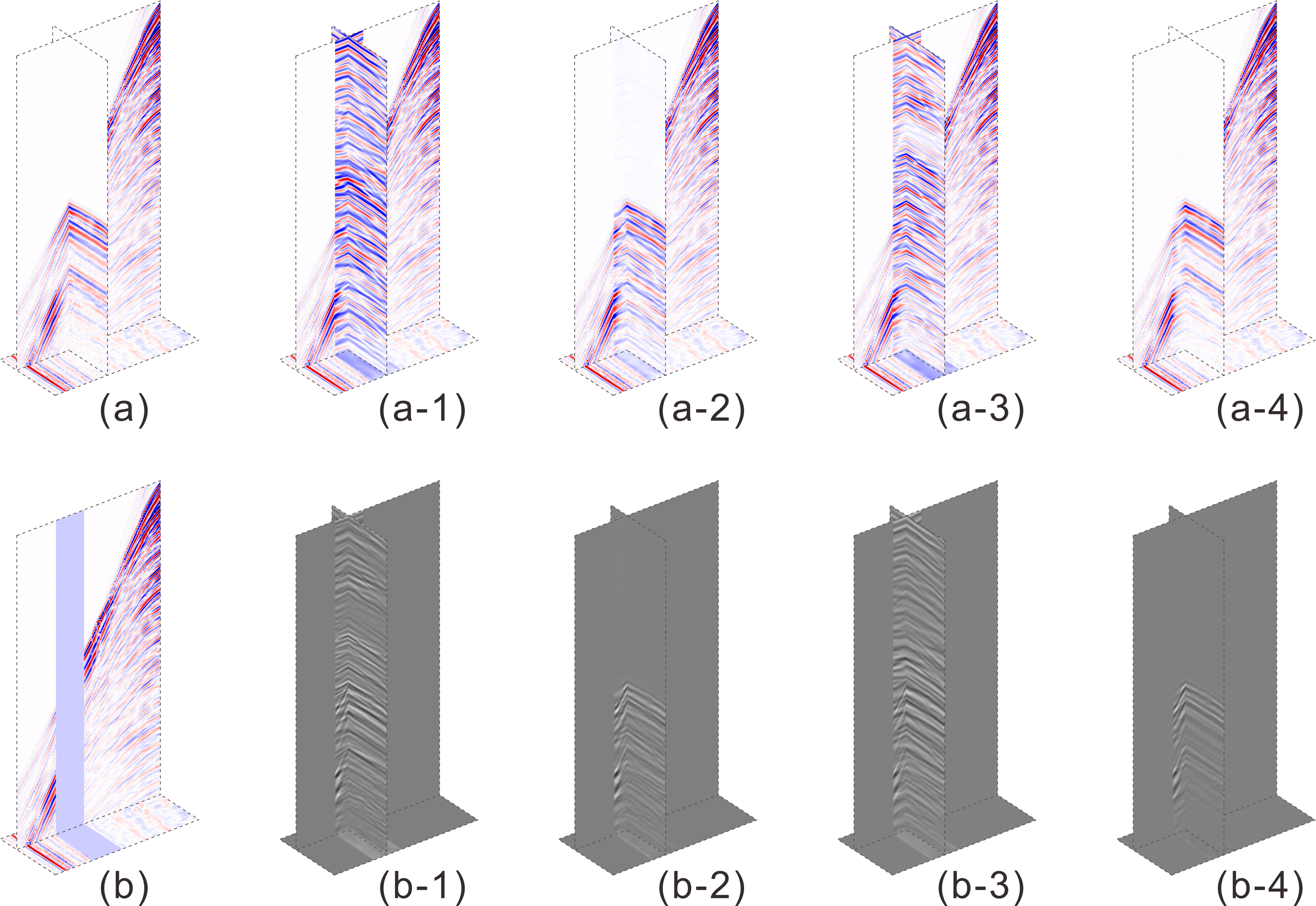}
		\caption{(a) Ground truth,(b) 50 continuous missing traces, (a-1) to (a-4) are the reconstruction results of Unet, conditional DPM, MDA GAN, and SeisFusion, respectively. (b-1) to (b-4) are the residuals of each reconstruction result relative to the ground truth.}
		\label{fanhua}
	\end{figure*}
	\section{Discussion}
	\subsection{Generalization}
	To explore the generalization ability of the proposed method, we conducted generalization experiments. The training set for the experiment consisted of synthetic data SEG C3. The model trained on SEG C3 was then directly used to reconstruct a more complex field dataset Mobil Avo Viking Graben Line 12. We intentionally removed 50 consecutive traces and set the traces to zero to represent the missing traces, which was used as the data to be reconstructed. The reconstruction results of the four models are shown in Fig. \ref{fanhua}. It can be observed that the generalization ability of convolution-based reconstruction methods, Unet and MDA GAN, is relatively weak. In the low amplitude portion of the upper triangle in the reconstructed data, they produced features similar to those of the SEG C3 dataset. This may be because convolution-based methods are a point-to-point reconstruction approach, learning the characteristics of the training set. When applied to untrained data with significant differences, this point-to-point reconstruction method can only mechanically reconstruct based on the characteristics of the training set. In contrast, the diffusion model-based reconstruction methods, DPM and our method, were able to reconstruct the basic features of the data. In the low amplitude portion of the upper triangle, the reconstructed data showed high consistency. This is largely due to the fact that the diffusion models learn the probability distribution of the target data and can provide an ideal coverage of the distribution, thus having stronger generalization ability. However, in the high amplitude portion of the lower triangle, there were more noticeable inconsistencies.
		\begin{figure*}[!t]
		\centering
		\includegraphics[width=0.8\textwidth]{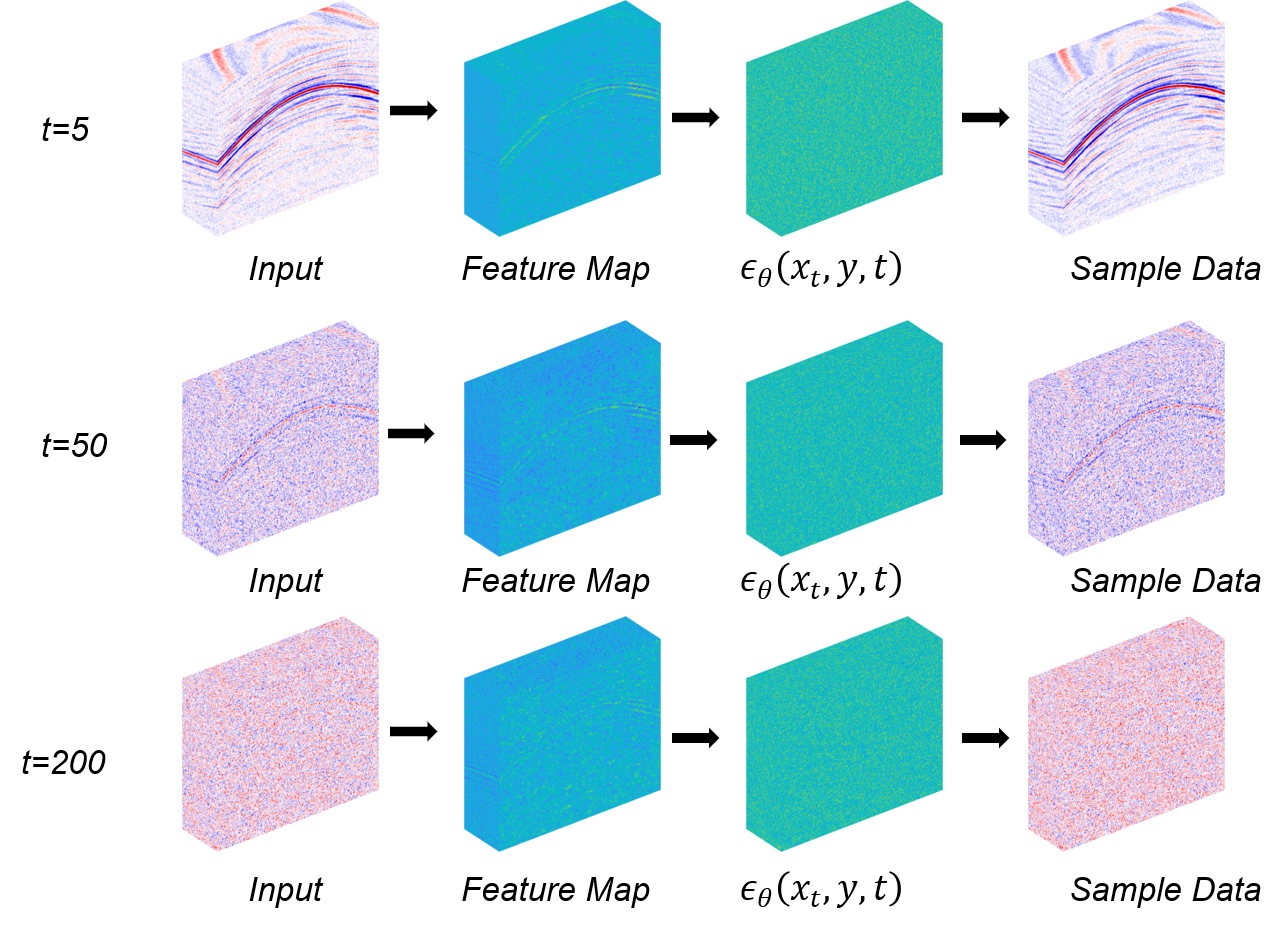}
		\caption{Feature maps extracted by the network from SEG C3 data with 50\% random missing at times t=5, t=50, and t=200, along with estimates for $\epsilon_\theta(x_t, y, t)$ and sampled data.}
		\label{feature}
	\end{figure*}
	We calculated four evaluation metrics in Table~\ref{tab:fanhua} to quantitatively assess the generalization ability of the four methods. The results show that the methods based on diffusion models have strong generalization ability. Moreover, thanks to guided sampling, our method outperforms DPM in terms of generalization ability and is already capable of preliminarily reconstructing more complex data.
	\begin{table}[!t]
		\caption{Comparison of Generalization Ability of Four Reconstruction Networks with 50 Consecutive Missing Traces in Field Dataset}\label{tab:fanhua}
		\centering
		\linespread{1.25} \selectfont
		\begin{tabular}{cccc}
			\hline
			& MSE       & SNR     & SSIM   \\ \hline
			UNet  & 2.0850e-3 & 26.8087 & 0.8930 \\
			Conditional DPM  & 7.1440e-4 & 30.4605 & 0.9390 \\
			MDA GAN & 1.5233e-3 & 28.1720 & 0.9063 \\
			SeisFusion & \pmb{4.7471e-4} & \pmb{31.2357} & \pmb{0.9490} \\ \hline
		\end{tabular}
	\end{table}
	\subsection{Fine-tuning}
	Although the model trained on the SEG C3 dataset captures some fundamental features of the complex field dataset Mobil Avo Viking Graben Line 12 and exhibits a degree of generalization, the reconstruction performance remains relatively low. Therefore, fine-tuning the model with a smaller amount of data can improve its reconstruction accuracy. To investigate the amount of data required for fine-tuning, we conducted an experiment using 5\%, 10\%, and 15\% of the field data for fine-tuning. For this experiment, 50 traces were continuously missing in Mobil Avo Viking Graben Line 12, with traces set to 0 to represent the missing data. The reconstruction results for models fine-tuned with different data amounts are shown in the Table~\ref{tab:fine}.
	\begin{table}[!t]
		\caption{Comparison of Models Fine-tuned With Different Amounts of Data of Four Reconstruction Networks with 50 Consecutive Missing Traces in Field Dataset}\label{tab:fine}
		\centering
		\linespread{1.25} \selectfont
		\begin{tabular}{cccc}
			\hline
			& MSE       & SNR     & SSIM   \\ \hline
			5\%  & 8.1423e-5 & 39.8665 & 0.9832 \\
			10\%  & 6.2259e-5 & 41.8856 & 0.9942 \\
			15\% & 6.5413e-5 & 41.6233 & 0.9948 \\\hline
		\end{tabular}
	\end{table}
	
	It can be observed that 5\% of the data is sufficient to fine-tune a high-performance model. Increasing the data amount to 10\% achieves the model's optimal state. Therefore, only 10\% of the data is needed for fine-tuning to obtain a high-precision reconstruction model.
	\subsection{Interpretability}
	With the rapid development of deep learning, there is increasing focus on how neural networks operate, and interpretable deep learning models have long been a research goal among scholars. Regarding the working principles of diffusion models, as elaborated in Fig. \ref{condition} and the section on Diffusion Model in our article, the neural network in diffusion models is used to estimate the noise scale of data at the current time step $ t $. The diffusion model enables theneural network to learn the noise scale corresponding to each time step $ t $ through a forward process. During sampling, starting from time step $ T $, the neural network estimates the noise scale of $ x_t $ and samples to obtain $ x_{t-1} $, iteratively obtaining $ x_0 $. Regarding how the neural network estimates the noise scale, we extract and plot feature maps of the neural network as well as the noise estimated by the network, as shown in Fig. \ref{feature}. We selected a case with 50\% random missing data reconstruction on the SEG C3 dataset and plotted features extracted by the network at three time steps: $ t=5 $, $ t=50 $, and $ t=200 $. Heatmaps generated from these plots show that when extracting features, the neural network pays more attention to the signal part of the input data. Even at $ t=200 $ when noise is substantial, signal features can still be extracted. This focused part gradually contributes to denoising to obtain clean data as desired. Based on the extracted feature maps and embedding of the current time step $ t $, the decoder of the neural network ultimately outputs an estimate of the noise $\epsilon_\theta(x_t, y, t)$. With $\epsilon_\theta(x_t, y, t)$ obtained, the mean and variance of the entire data distribution can be calculated. Subsequently, data can be sampled based on reparameterization techniques.

	\subsection{Time}
	In our guided sampling approach, an iterative process is introduced at each sampling step. Although this allows the diffusion model to use the data to guide and constrain the sampling process, it inevitably increases the sampling time, as the entire guided sampling process is longer compared to diffusion models that do not alter the sampling process. To reduce reconstruction time, DIM \cite{song2020denoising} can be used during reconstruction. The concept of DIM is compatible with our algorithm. When using the DIM algorithm, 50 sampling steps can achieve reconstruction quality equivalent to diffusion model, and fewer sampling steps, such as 20 or even 10 steps, can be used while maintaining acceptable reconstruction quality.
	
	\section{Conclusions}
	This paper proposes a 3D diffusion model with guided sampling and constraint incorporation for reconstructing complex 3D seismic data. Guided sampling utilizes input seismic data to guide the generation of reconstructed data by sampling from the given data during the sampling generation process. By incorporating constraints, uncertainties produced by the diffusion model's sampling process are successfully avoided, marking the first successful application of the diffusion model to 3D seismic data reconstruction. This method reconstructs data by learning the distribution of existing seismic data, effectively avoiding the performance degradation of traditional convolutional networks when faced the data to be reconstructed exhibits complex missing patterns, due to point-to-point learning reconstruction. Some ablation studies were conducted to validate the rationality of the hyperparameter settings used in the proposed method. Comparative experimental results on synthetic and filed datasets demonstrate that our proposed method yields more accurate interpolation results compared to other existing methods. The diffusion model itself, by learning the data distribution, possesses higher generative accuracy and generalization capability, allowing our network to generalize to more complex missing scenarios during inference.

	\normalem
	\bibliography{r}
	\begin{IEEEbiography}[{\includegraphics[width=1in,height=1.25in,clip,keepaspectratio]{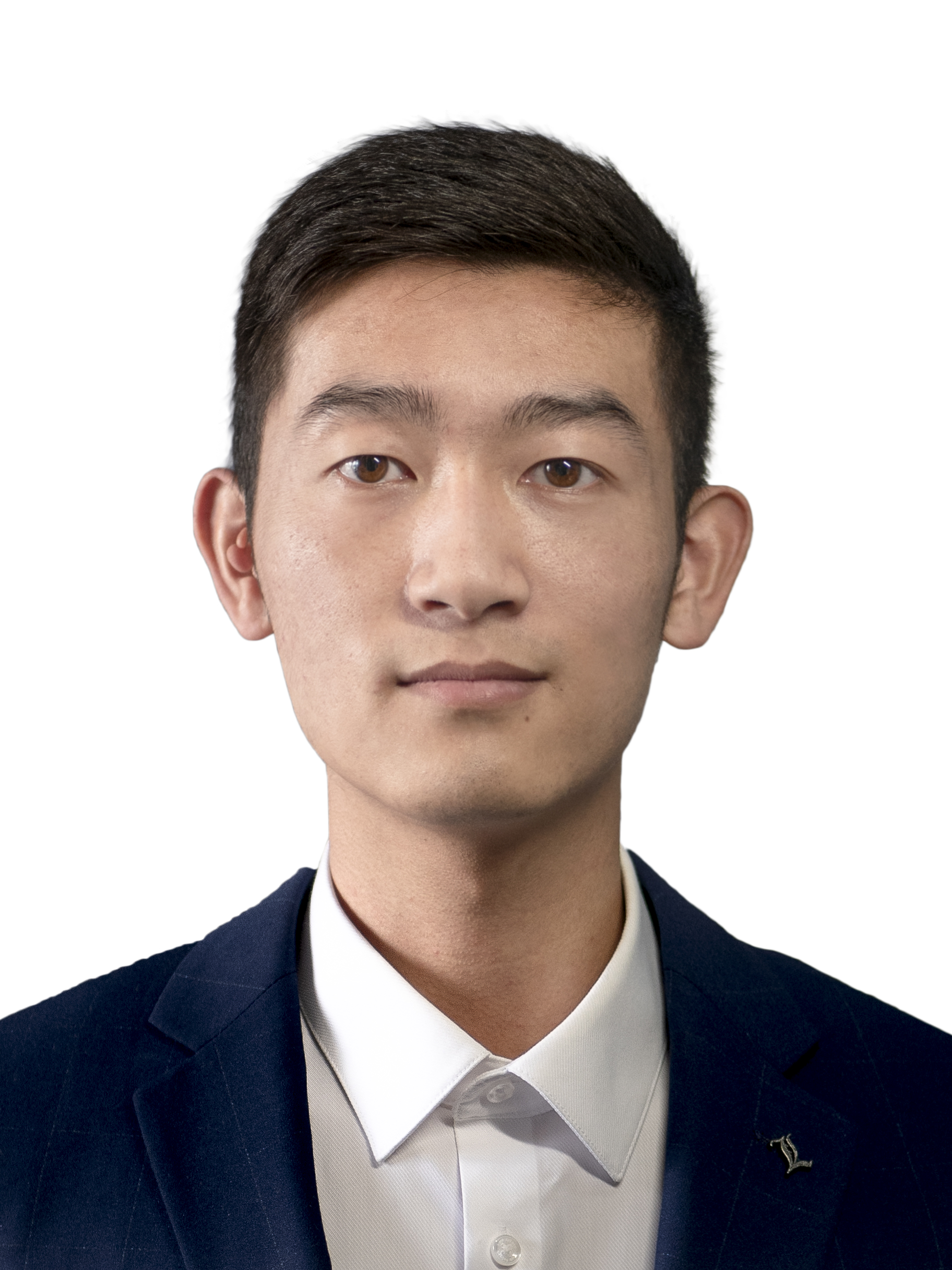}}]
		{Shuang Wang} received the B.S. degree in software engineering from the Chengdu University of Technology, Chengdu, China, in 2022. He is currently pursuing his Ph.D degree in Earth Exploration and Information Technology at Chengdu University of Technology, Chengdu, China.
		
		His research interests include applications of deep learning, computer vision, complex signal processing, intelligent geophysical data processing and modeling.
	\end{IEEEbiography}
	\begin{IEEEbiography}[{\includegraphics[width=1in,height=1.25in,clip,keepaspectratio]{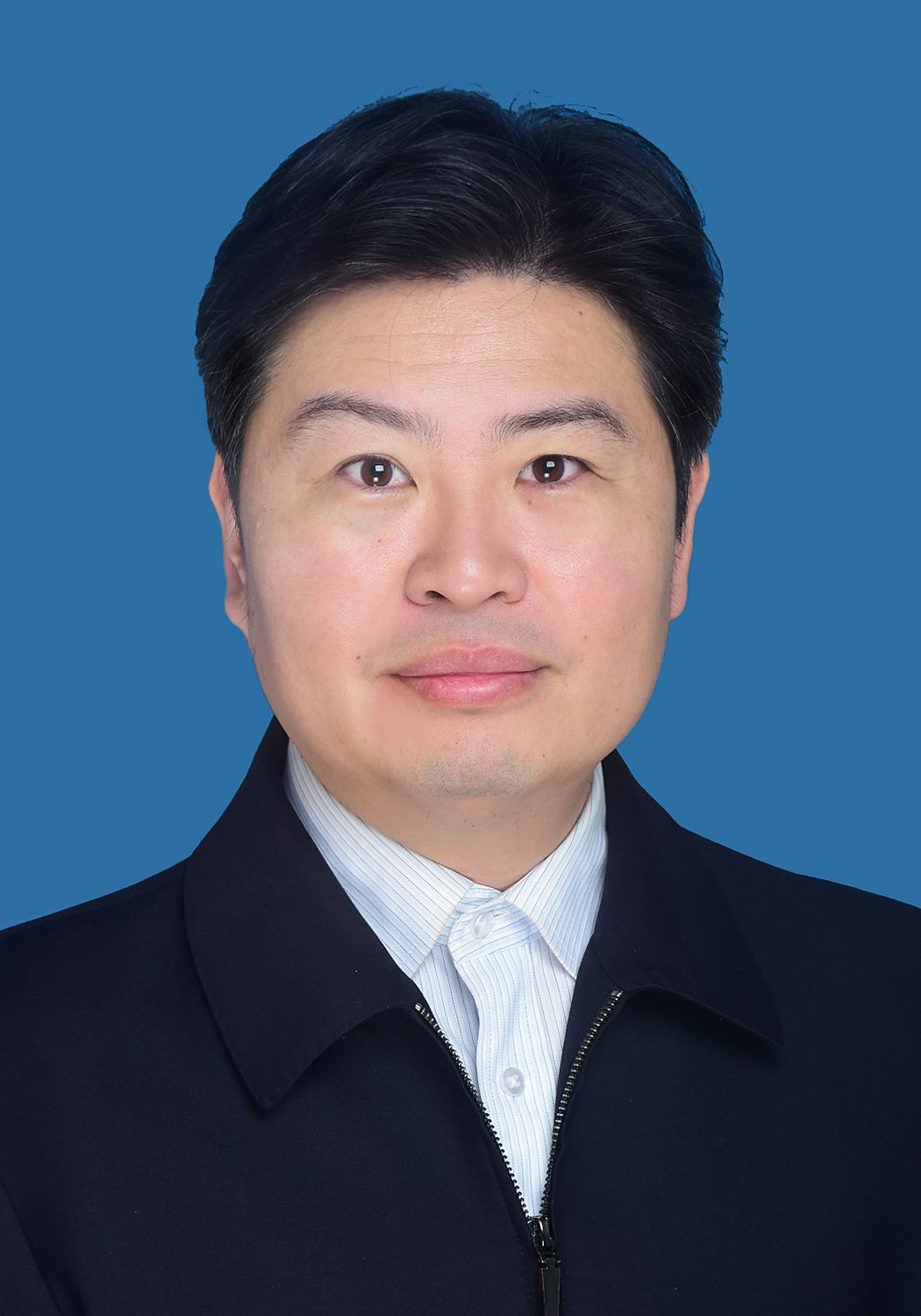}}]
		{Fei Deng}received the M.S. and Ph.D. degrees in Earth Exploration and Information Technology from the School of Information Engineering, Chengdu University of Technology, China, in 2004 and 2007, respectively. He has been with the College of Computer and Network Security, Chengdu University of Technology, where he is currently a Professor.
		
		His research interests include geophysical three-dimensional modeling methods, deep learning, and computer graphics.
	\end{IEEEbiography}
		\begin{IEEEbiography}[{\includegraphics[width=1in,height=1.25in,clip,keepaspectratio]{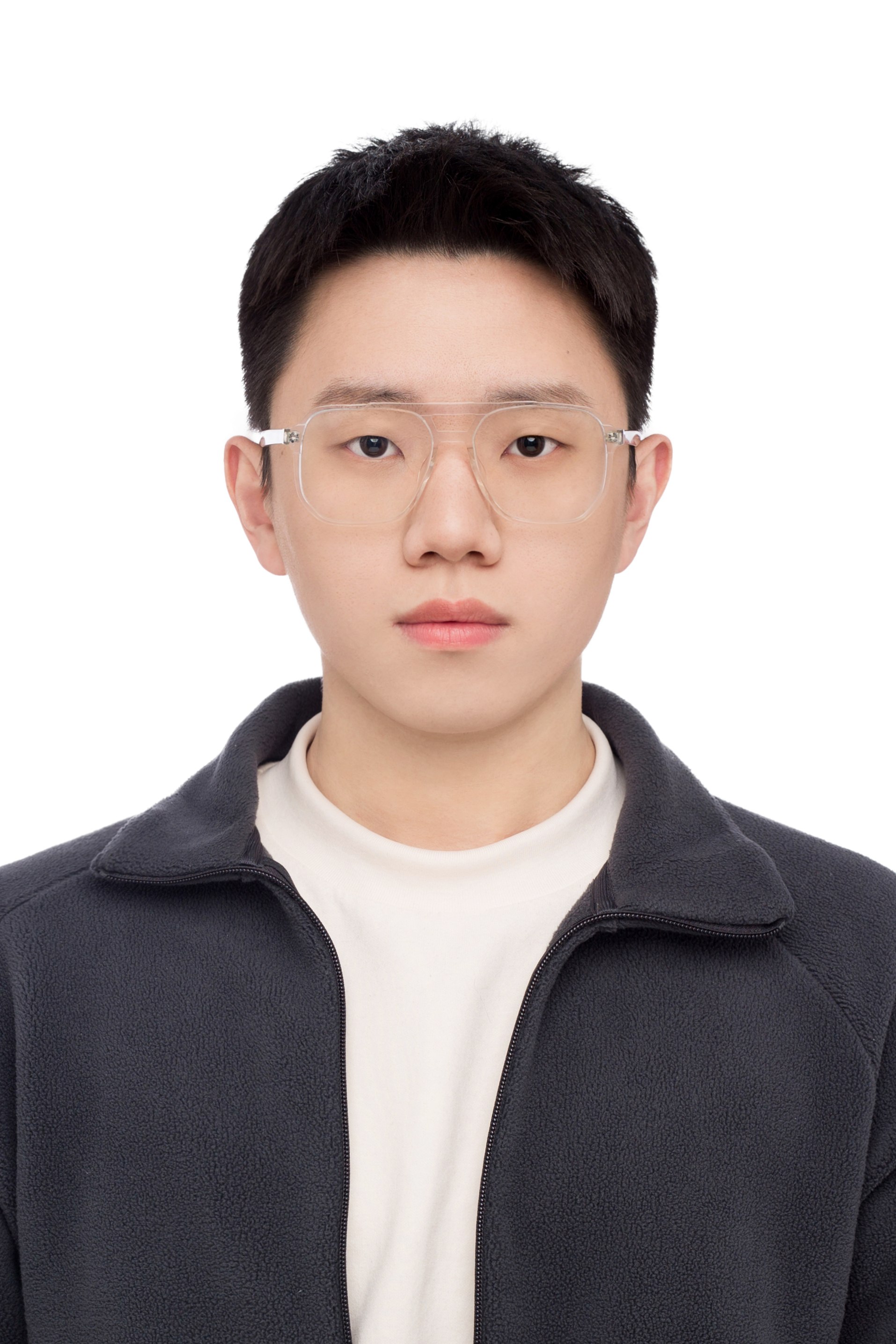}}]
		{Peifan Jiang}(Graduate Student Member, IEEE) received the M.S in Computer Technology from Chengdu University of Technology, Chengdu, China, in 2023. He is currently pursuing his Ph.D degree in Earth Exploration and Information Technology at Chengdu University of Technology, Chengdu, China.
		
		His research interests include applications of deep learning, intelligent geophysical data processing and modeling.
	\end{IEEEbiography}
	\begin{IEEEbiography}[{\includegraphics[width=1in,height=1.25in,clip,keepaspectratio]{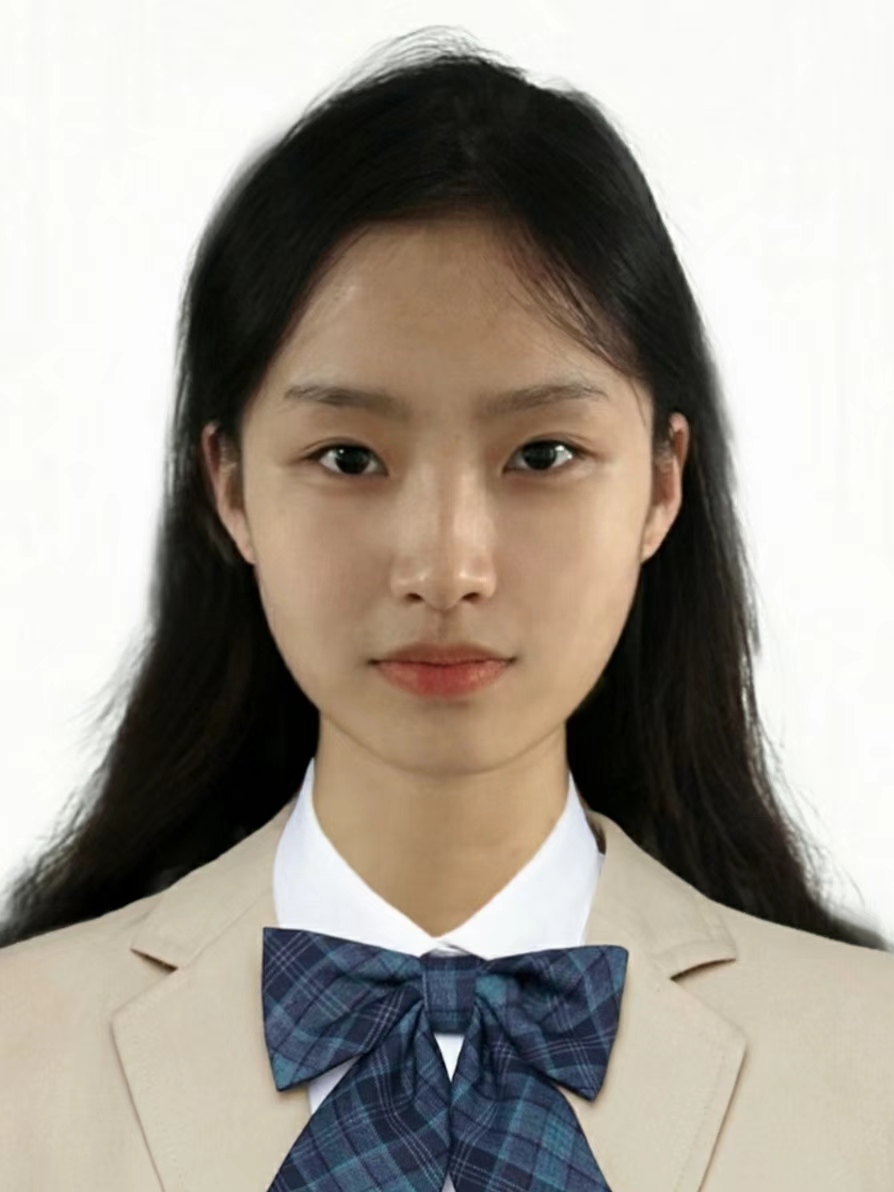}}]
		{Zishan Gong}received the B.S. degree in artificial intelligence from the Chengdu University of Technology, Chengdu, China, in 2024. She is currently pursuing the M.Sc. degree with Chengdu University of Technology, Chengdu, China, under the supervision of Prof. Fei. Deng. 
		
		His research interests include the computer vision and  deep learning.
	\end{IEEEbiography}
		\begin{IEEEbiography}[{\includegraphics[width=1in,height=1.25in,clip,keepaspectratio]{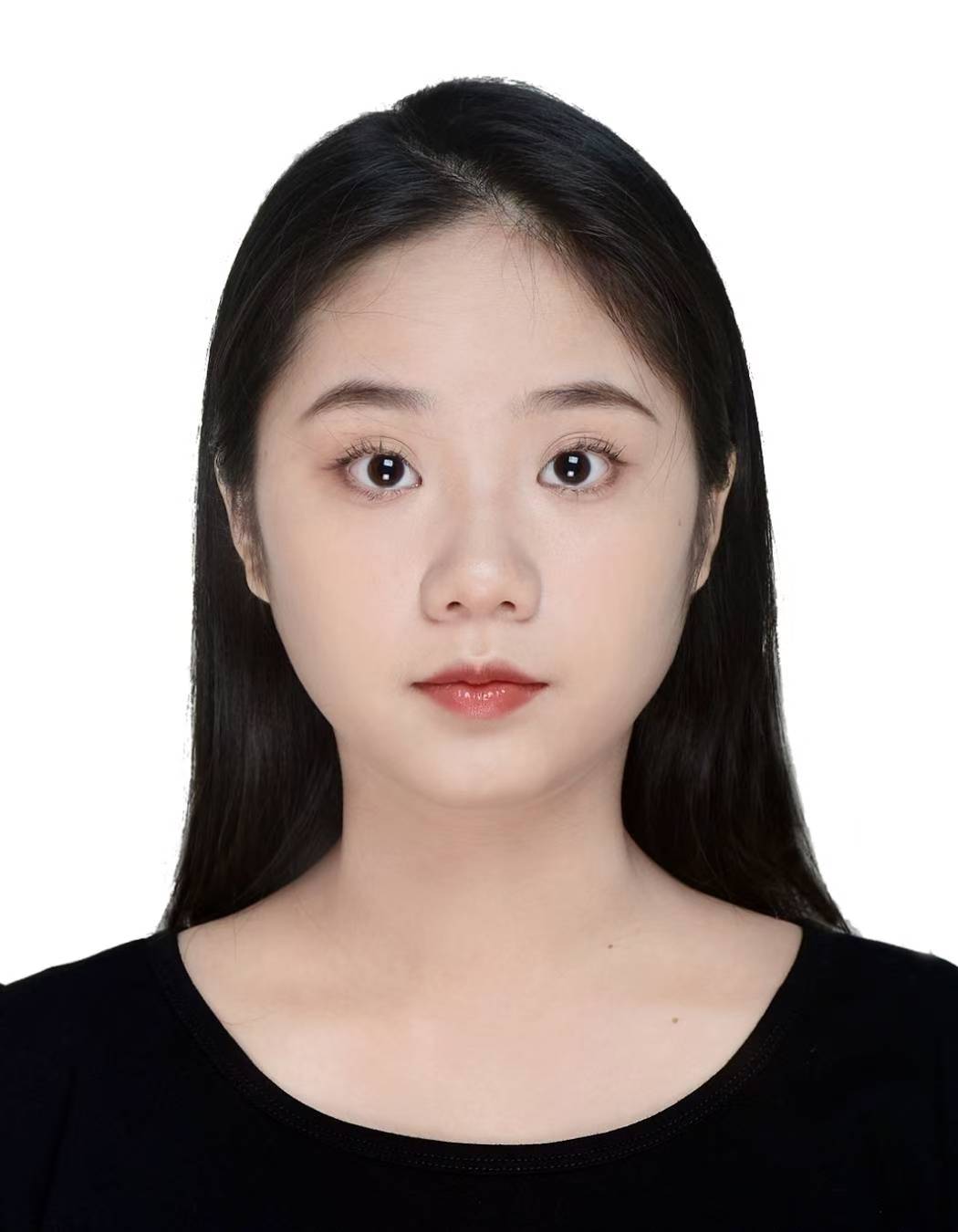}}]
		{Xiaolin Wei}received B.S. degree in digital Media Technology from Chengdu University of Technology in 2023. She is currently pursuing the M.Sc. degree at Chengdu University of Technology under the supervision of Prof.Fei. Dei. 
		
		Her research interests include the computer vision and deep learning.
	\end{IEEEbiography}
	\begin{IEEEbiography}[{\includegraphics[width=1in,height=1.25in,clip,keepaspectratio]{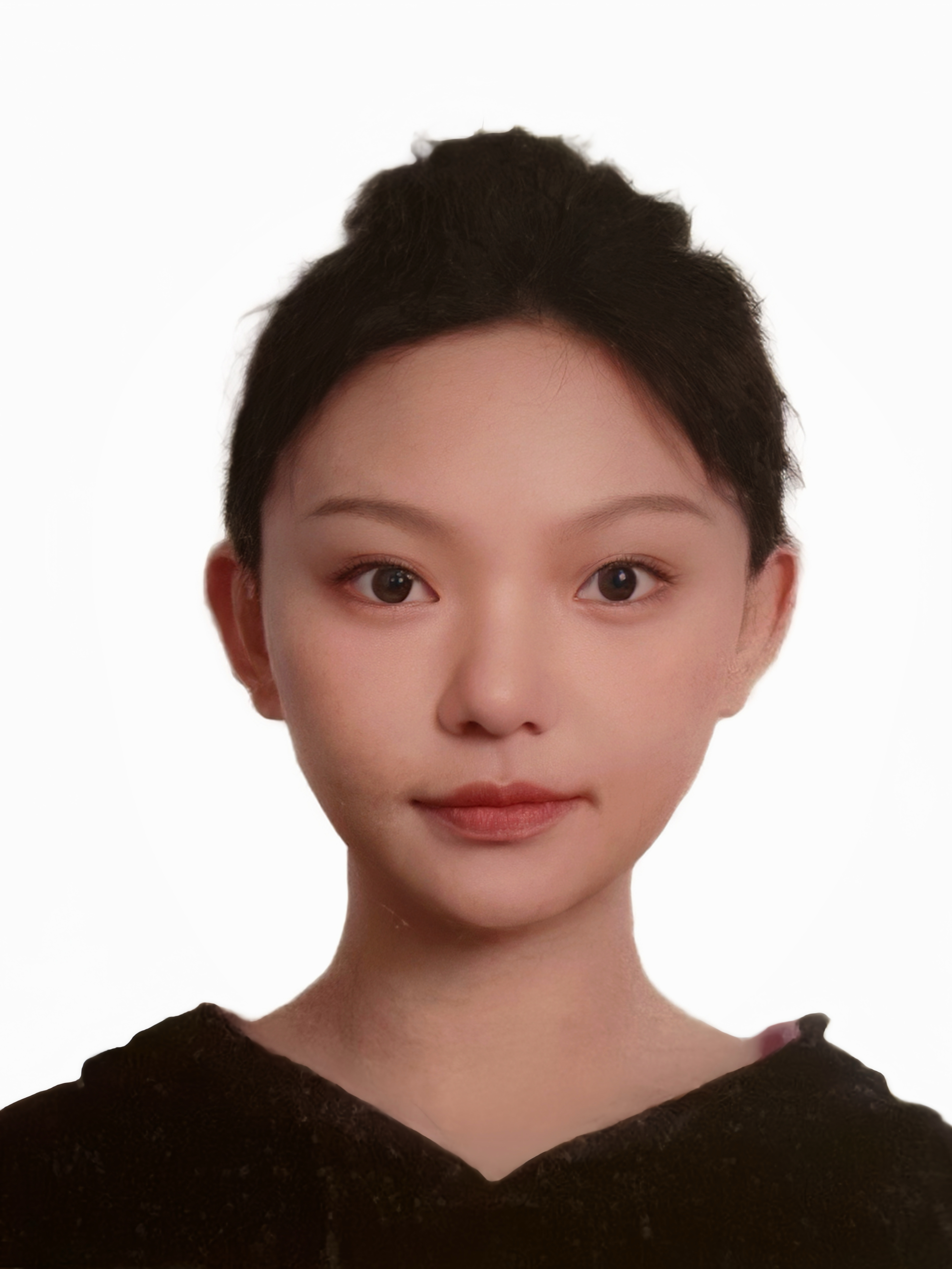}}]
		{Yuqing Wang}received the B.S. degree in software engineering from the Chengdu University of Technology, Chengdu, China, in 2023. He is currently pursuing the M.Sc. degree with Chengdu University of Technology, Chengdu, China, under the supervision of Prof. Fei. Deng.
		
		His research interests include the computer vision and  deep learning.
	\end{IEEEbiography}
	
	\bibliographystyle{IEEEtran}
	
\end{document}